\documentclass[10pt,twocolumn,letterpaper]{article}

\usepackage{cvpr}              

\usepackage{multirow}

\definecolor{cvprblue}{rgb}{0.21,0.49,0.74}
\usepackage[pagebackref,breaklinks,colorlinks,allcolors=cvprblue]{hyperref}
\usepackage{setspace}

\def\paperID{2035} 
\def\confName{CVPR}
\def\confYear{2025}

\title{PyTorchGeoNodes: Enabling Differentiable Shape Programs for 3D Shape Reconstruction}

\author{Sinisa Stekovic$^{1,2}$
\quad
Arslan Artykov$^{1}$
\quad
Stefan Ainetter$^{2}$
\quad
Mattia D'Urso$^{2}$
\quad
Friedrich Fraundorfer$^{2}$
\and
{\normalsize $^1$ LIGM, \'Ecole des Ponts et Chaussees, IP Paris, CNRS, France } \\ {\normalsize
$^2$ Inst. for Computer Graphics and Vision, Graz Univ. of Technology, Austria}
\and
{\tt\small Project page: \href{https://vevenom.github.io/pytorchgeonodes/}{vevenom.github.io/pytorchgeonodes}}
}

\usepackage{dsfont}

\newif\ifshowedits

\newcommand{\addeditor}[3]{%
  \definecolor{#1color}{rgb}{#3}
  \expandafter\newcommand\csname #1\endcsname[1]{
  \ifshowedits
    {\color{#1color} ##1}
  \else
    {##1}
  \fi
  }%
  \expandafter\newcommand\csname #1rmk\endcsname[1]{
  \ifshowedits
    {\color{#1color} {\bf [#2: ##1]}}
  \else
    {}
  \fi
  }%
}

\newcommand{\createtextvar}[1]{
  \expandafter\newcommand\csname #1\endcsname{%
  {\text{#1}}
}%
}
\newcommand{\textvars}[1]{\forcsvlist{\createtextvar}{#1}}

\newcommand{\calG}{{\cal G}}

\newcommand{\calL}{{\cal L}}

\newcommand{\calN}{{\cal N}}

\newcommand{\calP}{{\cal P}}

\newcommand{\calV}{{\cal V}}

\addeditor{sinisa}{SS}{1.0, 0.0, 0.0}
\addeditor{mattia}{MD}{0.8, 0.0, 0.8}
\addeditor{stefan}{SA}{0.0, 0.0, 0.8}
\addeditor{friedrich}{FF}{0.0, 0.0, 0.5}
\addeditor{vincent}{VL}{0, 0.8, 0}
\showeditsfalse

\begin{document}
\maketitle

\begin{abstract}
We propose PyTorchGeoNodes, a differentiable module for reconstructing 3D objects and their parameters from images using interpretable shape programs. Unlike traditional CAD model retrieval, shape programs allow reasoning about semantic parameters, editing, and a low memory footprint. Despite their potential, shape programs for 3D scene understanding have been largely overlooked. Our key contribution is enabling gradient-based optimization by parsing shape programs, or more precisely procedural models designed in Blender, into efficient PyTorch code. While there are many possible applications of our PyTochGeoNodes, we show that a combination of PyTorchGeoNodes with genetic algorithm is a method of choice to optimize both discrete and continuous shape program parameters for 3D reconstruction and understanding of 3D object parameters. Our modular framework can be further integrated with other reconstruction algorithms, and we demonstrate one such integration to enable procedural Gaussian splatting. Our experiments on the ScanNet dataset show that our method achieves accurate reconstructions while enabling, until now, unseen level of 3D scene understanding.
\end{abstract}

\newcommand{\param}[1]{\textit{\textsf{#1}}}

\def\moveinputvertically{1.0cm}

\begin{figure*}
    \centering
    \scalebox{0.9}{
    \begin{tabular}{cccc} 
    

    \multicolumn{4}{c}{\includegraphics[width=0.9\linewidth,trim=0 0 0 0.3cm,clip]{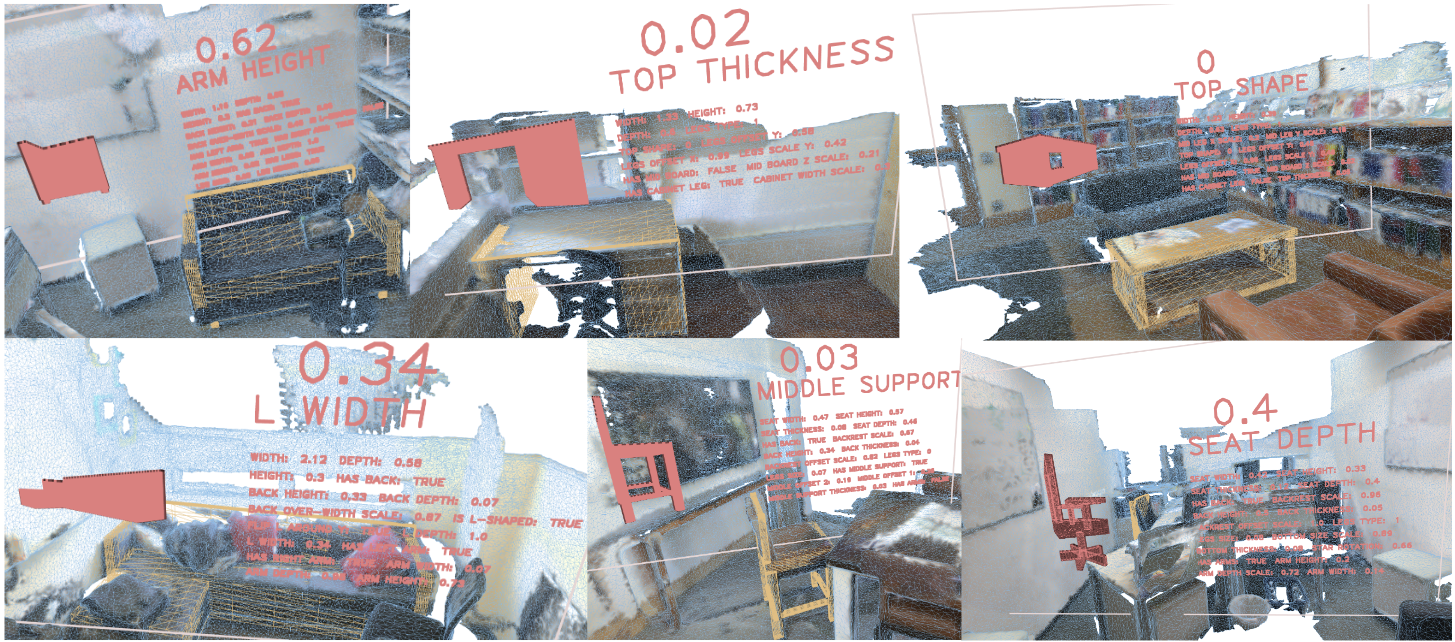}} \\
    
    \multicolumn{4}{c}{Jointly recovered geometry and parameters of target objects from RGB-D scans} \\

     & \includegraphics[trim={0.0cm 0.0cm 0.0cm 0cm},clip,height=3.0cm]{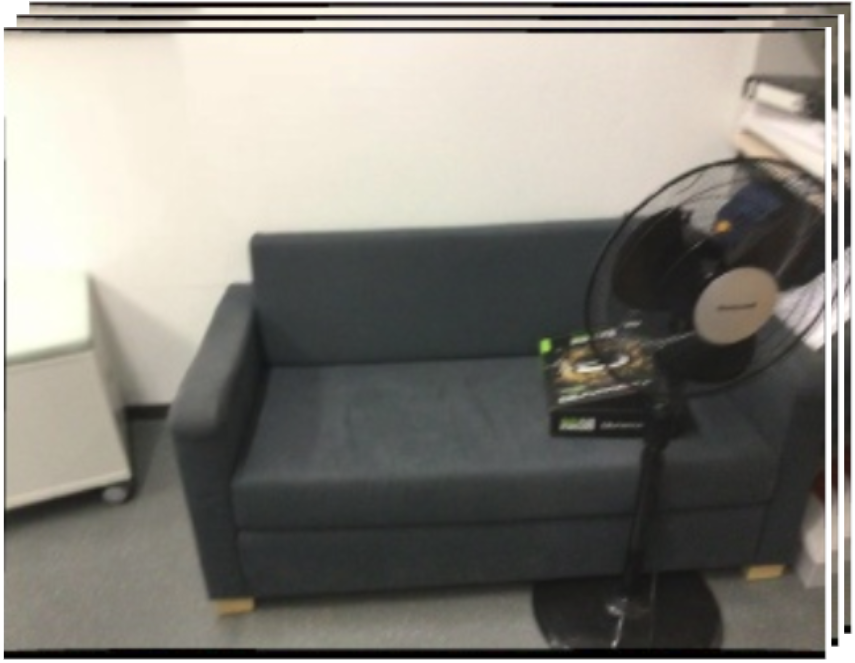} &

     \includegraphics[height=3.0cm]{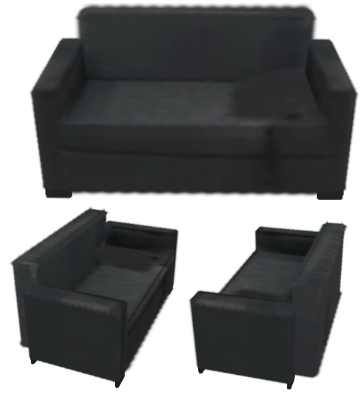} & 

     \includegraphics[height=3.0cm]{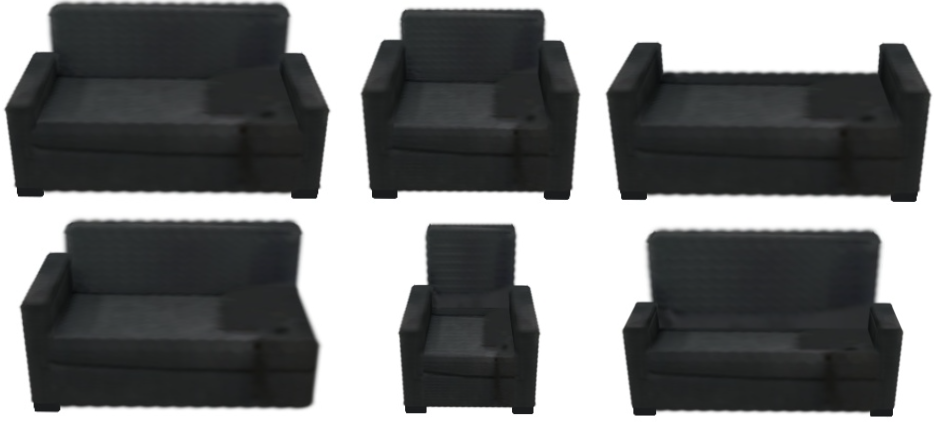} \\

         & Views of Target object & Recovered Gaussians & Procedural Gaussians editing \\
    \end{tabular}}
    \caption{Our PyTorchGeoNodes is a framework that enables differentiable shape programs for 3D reconstruction. Our semantically-rich reconstructions of 3D objects from RGB-D scans enable understanding of important metric parameters of individual parts for different objects. These parameters are physically bound to the 3D reconstruction as PyTorchGeoNodes enables flow of gradients from the generated shape to the parameters of the object.  Our PyTorchGeoNodes is modular, \eg we can use it to enable procedural modeling for Gaussian splatting. In the example, our method recovers Gaussians for the target sofa and 'clones' all details from the visible armrest to the occluded armrest in the scene. Our integration enables procedural editing of Gaussians and we can alter shape parameters to edit the reconstruction: \eg by 
 extending or shrinking width of the seat, removing backrest, left armrest, changing height of the backrest and depth of the seat \etc.}
    \label{fig:teaser}
    \vspace{-0.3cm}
\end{figure*}

\section{Introduction}

Various 3D object representations have already been considered for 3D scene understanding:  voxels~\cite{shin-cvpr18-pixelsvoxelsandviews,long2023wonder3d}, meshes~\cite{gkioxari2019mesh,wang2018pixel2mesh}, SDFs~\cite{park-cvpr19-deepsdf}, NeRFs~\cite{cao2023scenerf}, Gaussian Splatting~\cite{splatter}, exemplars from a shape database~\cite{aubry2014seeing}. SDFs, NeRFs, or Gaussian Splatting 
 generalize to shapes unseen during training but produce artifacts for example when parts of the object are occluded in scenes captured ``in the wild''. Their memory footprint can also become large.  Using exemplars guarantees that the retrieved shapes look good as they are  created by human designers, however the database may not contain shapes similar to the target object.

In this paper, we investigate the use of interpretable shape programs to represent objects. Shape programs are a form of procedural modeling~\cite{ebert2003texturing} introduced in \cite{tian2019learning} for representing 3D objects.
\vincent{While shape programs have already been considered for 3D scene understanding~\cite{simon-ijcv2011-shape,pearl2022geocode}, their use in this context remains largely unexplored. 
As shown in Figure~\ref{fig:teaser}, our method can be used to accurately retrieve geometry and 3D parameters of target objects, together with its texture even in presence of partial occlusions.}


Shape programs can be seen as computational graphs of parametrized geometric operations that generate various 3D shapes from an object category given input shape parameters, such as object dimensions as well as ``semantic'' properties like existence of armrest or backrest in case of chairs. Multiple similar parts, for example the legs of a chair or the shelves of a bookshelves---are obtained by first generating a single instance, duplicating it and applying geometric transformations such as rotations and translations as needed.

Thanks to this procedural modeling approach, shape programs generate shapes that remain appealing and that can fit many different instances. Alongside being interpretable, the shapes they generate are editable in an intuitive manner and have a very small memory footprint.
In fact, they are already a popular tool for building reconstruction~\cite{wonka2003instant,muller2006procedural,koehl-isprs15,vaienti2023} and the interest around shape programs for objects is growing~\cite{tian2019learning,jones2020shapeassembly,jones2021shapemod,pearl2022geocode}. Blender~\cite{blender2018} already provides support for them.  A research challenge targeting reconstruction of walls, doors, and windows using such structured languages was organized by Meta in 2024~\footnote{[{\href{https://eval.ai/web/challenges/challenge-page/2115/overview}{https://eval.ai/web/challenges/challenge-page/2115/overview (Accessed: November 2024)}}]}.

Recovering parameters of a shape program from an image is however challenging. Typically, a shape generated by a shape program depends on a number of parameters. Some parameters such as the width and height of the objects take continuous values but other parameters take only discrete values, \eg the number of shelves in a bookshelf.  

Optimizing on these parameters turns out to be tricky: For example, changing the number of shelves from 5 to 6 changes significantly the shape and appearance of a bookshelf. There have been a few attempts in the computer vision community to use procedural shapes~\cite{simon-ijcv2011-shape,pearl2022geocode}.  In~\cite{pearl2022geocode} the authors train a neural network to predict the parameters of a shape program. While this method shows impressive results using synthetic data, we show in Section~\ref{sec:eval_scannet} that on real data such an approach fails to generalize well, due to the lack of real world image data annotated with shape parameters. Using synthetically generated data to train such models also performs very badly: While shape programs capture very well the global structure of objects, they do not represent details. As a result, the model can hardly generalize from synthetic images to real ones. In comparison, our proposed method utilizes an objective function which is not based on learning and does not need training data, hence it is able to generalize well to real world scenarios.




We introduce three main contributions which enable the estimation of shape programs' parameters from images:
\begin{itemize}[leftmargin=0mm,itemindent=0.3cm]
\item Our first contribution is a `compiler` that generates PyTorch~\cite{paszke2019pytorch} code for a shape program, which can be defined in Blender for example. This code generates 3D shapes that are differentiable with respect to the continuous parameters and can therefore be used easily with 3D reconstruction algorithms to optimize the continuous parameters.
\item Second, we contribute a method to obtain efficiently and reliably all shape program parameters from RGB-D scans: While our compiler can be used to optimize over the continuous parameters, the challenge of estimating the discrete parameters remains. We propose to adapt methods based on genetic algorithms to our problem of jointly optimizing the continuous and the discrete parameters of a shape program. 
\item Our third contribution is an extension of shape programs to capture fine details and appearance of objects. We do this by augmenting our programs with Gaussians. Once we retrieved the shape 
 parameters, we initialize our \textit{procedural} Gaussians over the primitives and optimize them as in Gaussian splatting~\cite{kerbl3Dgaussians}. The key difference with regular Gaussian Splatting is that our Gaussians are constrained by the procedural model: The Gaussians sampled on 'cloned' base primitives are tied together. \eg as the legs of a chair are 'clones' of the same base primitives, the occluded part of a leg can still be recovered if it is visible for another leg.
\end{itemize}

We evaluate our method on both synthetic and real scenes. In addition to the usual metrics, we also evaluate how well we reconstruct the shape program continuous and discrete parameters. Because existing datasets are not annotated with the value of these parameters, we manually annotated $176$ objects.

In summary, our main contribution is a general tool to create PyTorch code from Blender shape programs. In addition, we provide a method for estimating the parameters of shape programs from views together with the poses of the objects. Our PyTorchGeoNodes is a modular tool and it can be extended to novel use cases in 3D scene understanding. The source code is available to the general public.

\section{Related Work}

Since the field of retrieving 3D models for objects from images is very large, we focus here only on methods that can provide a light CAD model designed for specific categories: 3D model retrieval methods and Shape Programs.

\subsection{3D Model Retrieval}

3D model retrieval aims to retrieve a CAD model from a database of objects, such as the ShapeNet dataset~\cite{chang2015shapenet}, that maximizes some scoring function which is typically based on an input 3D representation or a single or multiple input images.

Some methods approach CAD model retrieval as a classification problem~\cite{aubry2014seeing,mottaghi2015coarse}.  For a given input a neural network can be trained to predict a probability distribution over all models in the database. However, these direct approaches do not scale well to large object databases such as the ShapeNet database.  Another popular approach is computing a similarity metric between synthetic objects and real inputs.  There are several possible ways of computing such metrics. It is possible to render an image from a synthetic object, and then for a synthetic-real image pair, we can compute and compare image descriptors~\cite{aubry2015understanding,izadinia2017im2cad,massa2016deep} or encode images into embedding vector representations and calculate the similarity in the embedding space~\cite{kuo2020mask2cad}.


While such approaches have displayed promising results, one clear limitation is expressiveness of the database of objects. In fact, even with large databases of objects, in practice we often encounter objects in real scenes that are not represented in the database, or we might encounter modified or articulated objects. For example, we can modify seat height, arm rest and back support position of a work chair but the object in the database appears in the default setting which will introduce errors during retrieval. While it is possible to perform mesh refinement and deformation~\cite{botsch2004intuitive,sorkine2007rigid,jacobson2010mixed}, these approaches typically introduce artifacts and deformations in the reconstruction and the final object shape might not be representative of the actual object category after refinement.

\subsection{Shape Programs}

Procedural modelling is an attractive way of generating 3D shapes in computer graphics, and as such it is an interesting feature of different 3D modelling software. In general, procedural modelling is a family of techniques that generate 3D content by abiding to a set of rules. A variety of methods were proposed that combine procedural modeling with machine learning approaches, summarized in detail in~\cite{ritchie2023neurosymbolic}. For example, \cite{tian2019learning} introduced a procedural language that enables the generation of a variety of shapes in form of so called shape programs. In \cite{tian2019learning}, shape programs are written in Domain Specific Language~(DSL) for shapes with two types of statements. ``Draw'' statements describes shape primitive, such as cuboid or a cylinder, together with its geometric and semantic properties. ``For'' statements represent loops of sub-programs including parameters that indicate how the sub-program should be repeatedly executed. 

\cite{tian2019learning} also showed that it is possible to learn to infer shape programs that describe input 3D shapes. However, as it is based on basic geometric primitives, it becomes difficult to model more complex geometries, and resulting shapes can be non-representative of the target category. Other methods suffer from the same limitations~\cite{jones2020shapeassembly,jones2021shapemod}.  

GeoCode~\cite{pearl2022geocode} recently introduced hand-designed shape programs as a mapping between 3D shapes of a specific object category and human-interpretable parameter space. The language used for developing shape programs is based on the geometry node feature of Blender, an open-source modeling software, which, in contrast to~\cite{tian2019learning}, allows designing of more expressive shape programs that can represent and interpolate between a variety of shapes by simply modifying human-interpretable program's parameters. 

\begin{figure*}
    \centering
    \includegraphics[width=0.9\linewidth,trim=0 0 0.1cm 0,clip]{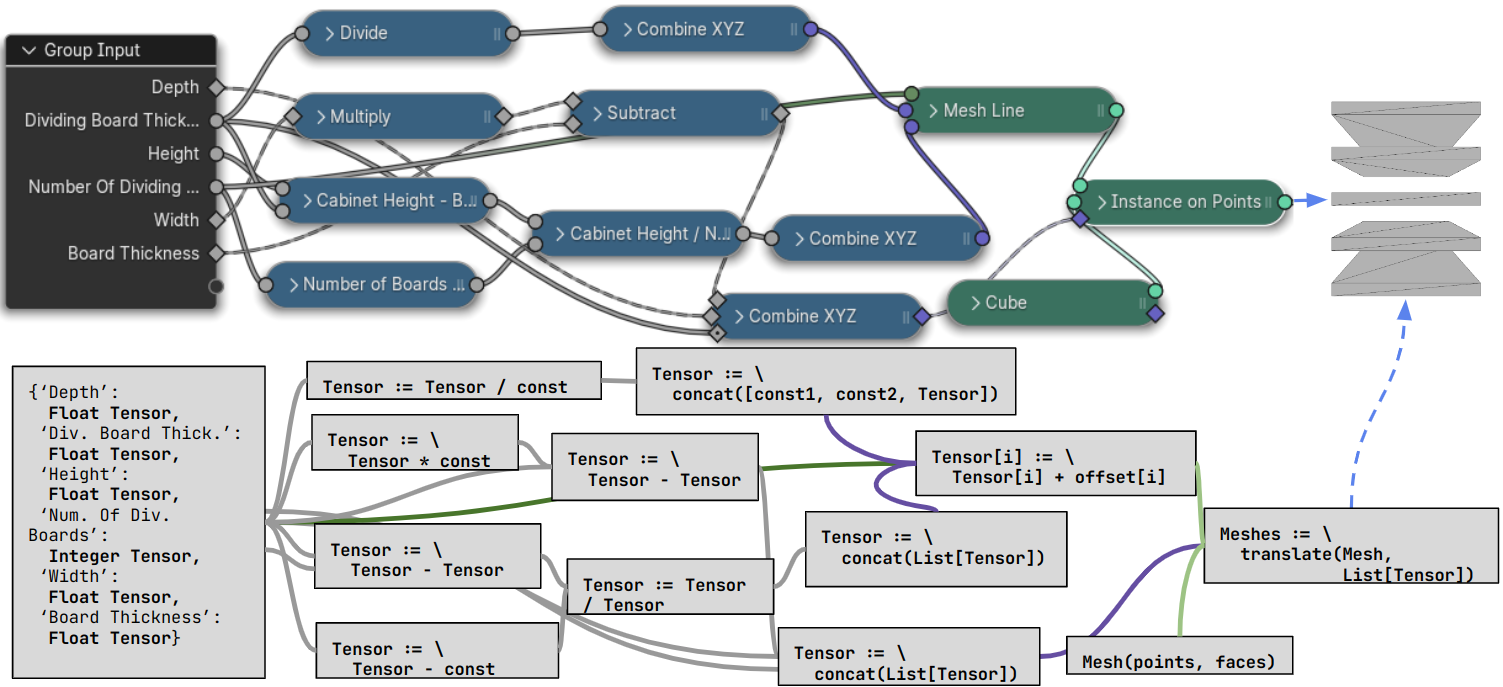}
    \caption{\textbf{A procedural model in Blender generating dividing boards of a cabinet.} The computational graph is designed and visualized in Blender using Geometry Nodes feature. Underneath, we show how we abstract the different nodes using PyTorch tensors and PyTorch3D meshes. The input node takes input parameters, here \{Width: 0.5, Dividing Board Thickness: 0.04, Height: 0.6, Number of Dividing Boards: 5, Board Thickness: 0.04\} and feeds them to a series of operations.  The blue nodes are arithmetic and concatenation nodes, which transform input parameters and feed the results to geometry nodes, in green. In this example, we generate a cuboid mesh and instantiate a line of points which generates the final geometry for dividing boards. In practice, this shape program is part of a larger shape program for modeling cabinets.
    \vincentrmk{this figure could go to the supp mat but dont forget to remove the refs to it in the text}
    }
    \label{fig:geonodes}
\end{figure*}

Some nodes define a shape primitive or pre-designed more complex shapes, \eg, chair seat or chair leg. Other nodes represent geometric transformations on individual shapes, or how two shapes should be connected, e.g., legs should always be connected to the bottom of the seat. There are nodes representing mathematical operations such as addition and multiplication, and there are nodes that are used to replicate specific shapes, e.g., in case of chair legs we can add a replicator node to create four instances of a chair leg of the same shape. There are also nodes that serve as control statements, e.g., it is possible to switch between a four-legged chair and a one-legged chair. A pre-defined set of input program's parameters allow non-expert user to easily control shape-generation process. \cite{pearl2022geocode} also showed that it is possible to learn the mapping from input point clouds of 3D shapes, or a sketch, to program's parameters. 


\section{PyTorchGeoNodes}



In this section, we discuss shape programs and how they can be beneficial for tasks in 3D scene understanding. We then describe our framework for enabling efficient gradient-based optimization for shape programs and, in addition, allowing to translate shape programs designed in Blender to PyTorch code to enable differentiable optimization. In Section~\ref{sec:search}, we explain how our framework can be applied for reconstructing 3D objects from images.


We consider shape programs as defined in Blender, i.e.,  as a computational graph that computes 3D shapes given a number of input parameters. Such graphs contain nodes of different types; in our framework, we currently consider the following node types:
\begin{itemize}
\item An \textit{'Input' node} contains input parameters of the shape programs and their value ranges. We consider floating, integer, and boolean values;
\item A \textit{'Math' node} performs mathematical operations on its  input~(addition, subtraction, multiplication, division, and comparison);
\item A \textit{'Switch' node} switches between two inputs conditioned on a boolean input;
\item A \textit{'Combine' node} concatenates three input values into a 3D vector;
\item A \textit{'Primitive' node} creates a geometric primitive, e.g., a cuboid or a cylinder;
\item A \textit{'Transform' node} applies scaling, rotation, and translation to the input geometry;
\item \textit{'Mesh Line' node} creates a tensor of 3D points on a line from input vectors that define the start location and the offset of the line, and an integer  that defines the number of points to be generated;
\item A \textit{'Points on instances' node} clones input geometry at given input 3D points;  
\item A \textit{'Join geometry' node} merges geometries;
\item An \textit{'Output geometry node'} generates the final mesh for the shape.
\end{itemize}
Note that for example, the Transform node depends on continuous, the Points on instances node on integer, and the Switch node on boolean.



Our framework provides differentiable computational nodes that reimplement the functionalities of the individual geometry nodes in Blender as we show in Figure~\ref{fig:geonodes}. More exactly, for every node type in Blender, we implement a corresponding node type with the same functionalities using PyTorch, or PyTorch3D~\cite{ravi2020pytorch3d} in case of geometric operations. In practice, we designed 
shape programs directly using the Blender interface. Given a shape program created in Blender, our parser generates a differentiable PyTorchGeoNodes code.  We can then use this code in PyTorch in an optimization procedure where we fit shape parameters to generate a shape consistent with the input scene.


In the supplementary material, we detail our design choices and technical details which make our implementation computationally efficient. 




\begin{figure}
    \centering
    \begin{tabular}{c}
             \includegraphics[width=0.95\linewidth]{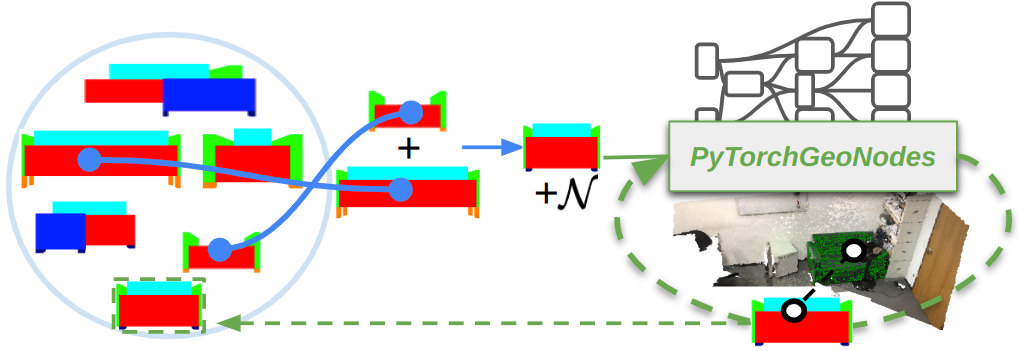} \\
    \end{tabular}
    \caption{\textbf{Fitting shape parameters with PyTorchGeoNodes.} At every iteration, our genetic algorithm generates indiviudals of shape parameters and object poses. We parse 3D shapes from individual parameters using PyTorchGeoNodes and perform shape parameter optimization based on loss that evaluates how well the shape fits the input scene. In the next iteration, the optimized set of individuals is used to update  the pool of individuals based on selection of the fittest principle.}
    \label{fig:pipeline}
\end{figure}

\section{Search Algorithm}
\label{sec:search}

\textvars{CD,PC,NN,G,Fl,LL,Sh,Visits, Points}

We leverage our PyTorchGeoNodes framework to retrieve shape parameters and geometry of a target object in an RGB-D scan. We look for shape parameters $\calP^*$ by minimizing an objective function that encourages the shape to fit the point cloud reconstructed from the images while lying on the floor: 
\begin{align}
 \calL(\calP) = & \lambda_\CD \calL_\CD(\PC, \Sh(\calP)) +
\lambda_\G \calL_\G(\G,\Sh(\calP))  + \nonumber \\
 & \lambda_\Fl \calL_\Fl(\PC, \Sh(\calP))) \> .
\label{eq:loss}
\end{align}
$\calL_\CD(\PC, \Sh(\calP))$ measures the single-direction chamfer distance between $\PC$, the point cloud reconstructed from the depth maps and object masks, and $\Sh(\calP)$, the shape generated by the shape program. We follow pre-processing from~\cite{ainetter2023hocsearch} which relies on sparse point cloud instance segmentation of the object and use a segmentation model~\cite{kirillov2023segment} on images to obtain the object masks. We then back-project these masks to obtain a dense target point cloud.
$\calL_\G(\G,\Sh(\calP))$ penalizes occlusions in the scene caused by $\Sh(\calP)$. $\G$ is an occlusion grid obtained by densely sampling 3D points along rays between the camera and surfaces for different views:
\begin{equation}
    \calL_\G(\G,\Sh(\calP)) = \sum_{p \in \Sh(\calP)}\begin{cases}
        \stackrel{\rightarrow}{\CD} (p, \PC) & \text{if } \G(p) \\ 
        0 & \text{otherwise},
    \end{cases}
\end{equation}
where $\stackrel{\rightarrow}{\CD} (p, \PC)$ is the distance of point $p$ to \PC. $\G(p) = \stackrel{\rightarrow}{\CD} (p, \PC) < \omega$ determines whether $p$ is on grid $\G$. We set threshold $\omega=5cm$ in our experiments. Since $\G$ is generated using input views for the scene, this loss does not penalize points that are occluded in the scene, similar to render-and-compare objectives. 
%
%
%
%
%
The term $\calL_\Fl$ simply enforces the objects to stay on the floor plane:
 \begin{align}
     \calL_\Fl(\PC, \Sh(\calP))) = |\Fl(\PC) - \Fl(\Sh(\calP))|_2 \> ,
 \end{align}
where $\Fl(\PC)$ returns the height of the lowest point in $\PC$, and $\Fl(\Sh(\calP))$ similarly determines the lowest point of shape $\Sh(\calP)$. We set weights of individual loss terms to $\lambda_\CD=1, \lambda_\G=0.5, \lambda_\Fl=0.01$ in our experiments. 
Parameters $\calP$ include the object pose in addition to the shape program parameters. We initialize the translation based on an initial estimate of the object center which is refined during search. We constrain the rotation to be around the vertical axis only.



We use our PyTorchGeoNodes framework to compute $\Sh(\calP)$. Thanks to this framework, PyTorch can compute the derivatives of $\Sh(\calP)$ and thus the derivatives of $\calL$ with respect to the continuous parameters in $\calP$. However, some parameters in $\calP$ are discrete, i.e., integers or booleans. Therefore, we cannot optimize $\calL$ using a simple gradient descent. Also, $\calL$ has no special form that we can exploit to optimize it. 

To solve this problem, we propose to resort to 
genetic algorithms to optimize the parameters and combine it with gradient descent to further refine the continuous parameters. 

\subsection{Shape Parameter Search}

A genetic algorithm is a stochastic optimization algorithm that iteratively applies random manipulations of high-dimensional search parameters to identify the most promising complex search subspaces, subject to an objective function. In contrast to gradient-based techniques, it can optimize discrete and continuous values and the objective function can be very general and non-differentiable.

The algorithm starts by randomly generating instances of parameter values $\calP$, \textit{individuals}, that form an initial set called \textit{population} of search parameters. Every iteration has three phases: \textit{crossover} and \textit{mutation} for exploring search subspaces, and \textit{selection} for exploiting the most promising subspaces. Crossover randomly selects pairs of individuals, \textit{parents}, and mixes them by combining the individual parameters to create a new \textit{offspring}. Therefore, an offspring lies in the search subspace at intersection of parents' search spaces. Mutation then randomly alters offsprings' parameters to further explore this subspace. Selection evaluates the quality of the population, and based on \textit{survival of the fittest} strategy, it takes the best performing individuals to form the population for the next iteration. Further details and hyper-parameters are given in the supplementary material.

Our adaptation of genetic algorithms to our problem is described in details in the supplementary material. The main points are given below. We generate an initial population by randomly generating instances of all program's parameters $\calP$ representing individuals in the population. We draw random values from the empirical distributions of the parameters. The crossover operation 
generates offspring $P_{1,2}$ from a pair of parents $\calP_1$, $\calP_2$.
Mutation varies depending on the parameter type. For discrete parameter $d \in \calP_D$, we adapt its value $v_d$:
\begin{equation}
    v_d \leftarrow 
    \begin{cases} 
        v_s \in_R \calV_{d} &\quad \text{with probability } P_m \\
        v_d &\quad \text{otherwise},
    \end{cases}
\end{equation}
where operator $\in_R$ randomly selects value $v$ from the set of valid discrete values $\calV_p$. $P_m$ is a hyper-parameter that defines the probability of mutating parameter values. In case of a continuous parameter $c \in \calP_C$, we adapt its value $v_c$ as: 
\begin{equation}
     v_c \leftarrow 
    \begin{cases} 
        v_s \in_R \calV_{c} + \calN(0, \sigma) & \text{with prob. } P_m \cdot P_m \\
        v_c + \calN(0, \sigma)  & \text{with prob. } P_m \\ 
        v_c & \text{otherwise},
    \end{cases} 
\end{equation}
where $\calV_c$ is a discretisation of values for parameter $c$ that we further discuss in the supplementary. $\calN(0, \sigma)$ is normal distribution with mean value set to $0$ and standard deviation $\sigma$. Finally, we evaluate offsprings, append them to the current population and we select the best individuals based on a score function, in our case negative of objective in Equation~\ref{eq:loss}, to form the population for the next iteration.

\textbf{Optimization step.} 
Using PyTorchGeoNodes, during the selection phase of every iteration, we perform a gradient-based optimization to further refine values of continuous shape parameters of individuals before selecting the population for next iteration. After a specified number of iterations the search concludes and we return shape parameters of the best performing individual.

\subsection{Integrating Gaussians Splats}

Shape programs explicitly capture symmetries and similarities between repeated structures, while this aspect is lacking in many representations for 3D reconstruction. On the other hand, earlier shape programs do not capture well details and appearance. 


In this section, we show we enable procedural Gaussian splats by integrating them within shape programs and our PyTorchGeoNodes framework. This allows us to capture fine details of objects while keeping the advantages of shape programs.  First, we explain how we augment the primitives with 'Gaussian primitives' while keeping the graph of PyTorchGeoNodes differentiable, and then, we introduce our render-and-compare objective function. 

\begin{figure}
    \centering
    \scalebox{0.8}{
    \begin{tabular}{ccc}
         \includegraphics[trim={0.3cm 0cm 0 0.3cm},clip,width=0.4\linewidth]{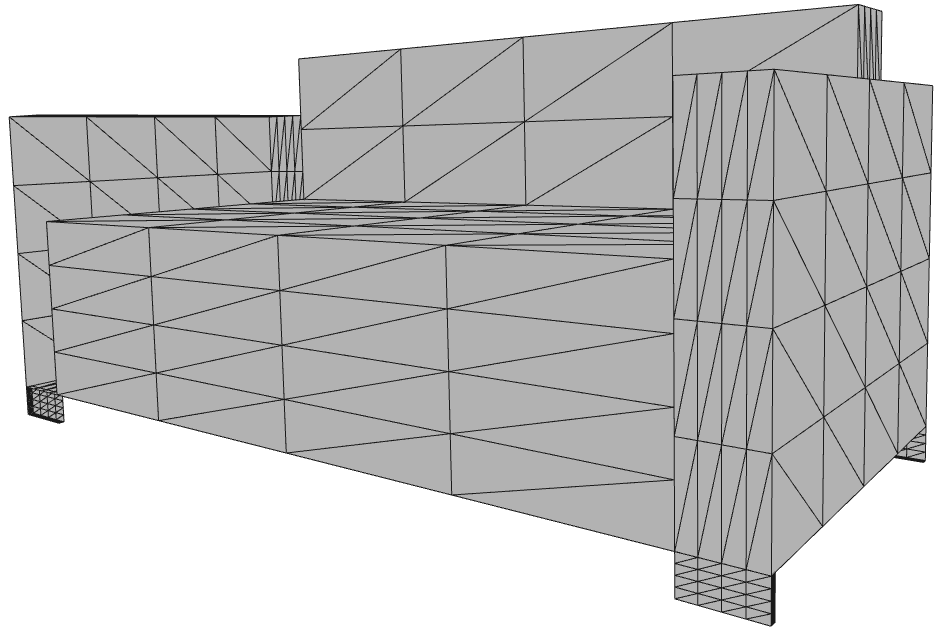} & \includegraphics[trim={0.3cm 1cm 0 0.0},clip,width=0.5\linewidth]{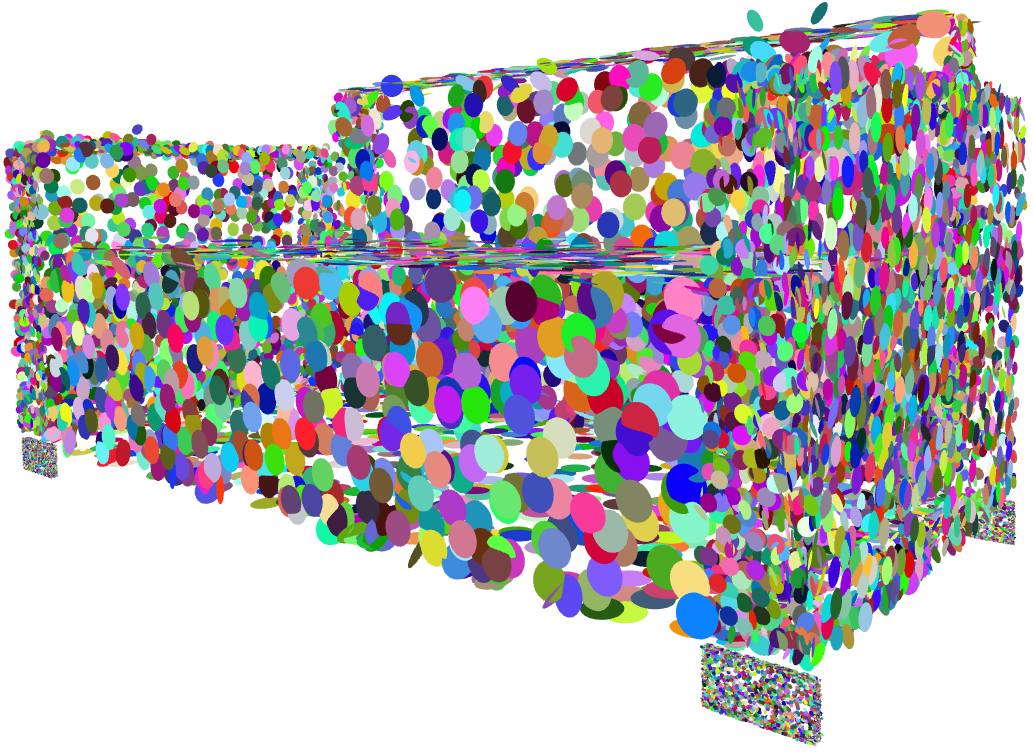}  \\
         Original Mesh & Augmented with Gaussians \\ 
    \end{tabular}}
    \vspace{-0.2cm}
    \caption{We integrate Gaussians directly into shape programs and PyTorchGeoNodes, to have the advantages of both procedural models and Gaussian splatting.}
    \label{fig:gaussians_method}
    \vspace{-0.4cm}
\end{figure}

\textbf{Extending PyTorchGeoNodes with Gaussian primitives.} To enable integration of our procedural graph to support Gaussian splats, we extend the nodes creating mesh primitives to also create trainable Gaussian parameters. We subsample the primitives meshes into many triangles and create Gaussians from these triangles. We initialize the Gaussians means, scales, and quaternions to fit the mesh triangles as we show in Figure~\ref{fig:gaussians_method}. Their colors are initialized based on the closest 3D point in the scan. Their opacities are fixed to a high value. As with other primitives, the graph transforms the Gaussians to form a complete set of Gaussians, some of them sharing parameters due to the cloning in the graph. 

\textbf{Optimizing the Gaussians.} 
We noticed that it is more robust to, rather than jointly, first estimate the shape parameters and then optimize the Gaussians. Given a set of optimal shape parameters $\calP$, we  minimize the render-and-compare loss as in other Gaussian splatting works, by rendering the Gaussians $\calG(\calP)$ into scene views~\cite{wu2024recent,kerbl3Dgaussians,guedon2024sugar}:
\begin{equation}
    \calL_\text{RGB} + \lambda_1\calL_\text{D} + \lambda_2\calL_\calP + \calL_\text{reg} \> ,
\end{equation}
where $\calL_\text{RGB}$ and $\calL_\text{D}$ are standard losses on color and depth consistency used in Gaussian splatting literature~\cite{wu2024recent,chung2024depth,turkulainen2024dn}. 

$\calL_{\calP}=\CD(\Sh(\calP), \calG(\calP))$ measures the consistency between the shape parameters and the Gaussians by relying on the chamfer distance. $\calL_\text{reg}$ is a regularizer term:
\begin{align}
        \calL_\text{reg} =& \lambda_3|M(\calG(\calP))|_2 + \lambda_4|S(\calG(\calP))|_2 + \\ 
    &\lambda_5 \sum_{G \in \calG(\calP)} \text{Var}(C(G)| , 
\end{align}
to enforce mean offsets $M(\calG(\calP))$ and scales offsets $S(\calG(\calP))$ of Gaussians to be small and the colors of the Gaussians on the same primitive $C(G)$ to remain close to each other. We set $\lambda_1=0.1,\lambda_2=0.1,\lambda_3=100,\lambda_4=10.00,\lambda_5=1$ in our experiments. In addition, we observed that optimizing object Gaussians and background Gaussians jointly at the same time, with loss for background locations set to $\lambda_{bgd} (\calL_\text{RGB} + \lambda_1\calL_\text{D})$ and $\lambda_{bgd}=0.5$ ensures that object Gaussians remain consistent around edges. 

{
\setstretch{0.95} 
\section{Validation}
\label{sec:evaluation}

We validate our PyTorchGeoNodes framework and our optimization method on three common furniture categories: Sofa, with $18$ parameters, Chair with $20$, and Table, with $10$, for which we created shape programs in Blender. We designed our shape programs to capture interpretable object structures that are relevant for objects in ScanNet, allowing unseen-before reasoning about high- level object parameters. More details are provided in the supplementary material.
%
 
Our method jointly estimates the pose and shape parameters of the object. 
The orientation is defined as the rotation around the up-axis as done in to~\cite{ainetter2023hocsearch}, which is sufficient for indoor scenes.

We validate our method on object shape reconstruction for real world scenes from the ScanNet dataset~\cite{dai2017scannet}. ScanNet is a challenging dataset of indoor environment RGB-D scans: Furniture is 
 only partially visible and highly occluded, and depth maps are incomplete due to missing sensor values---this happens for dark and reflective materials, for example. We manually annotated $176$ objects for Sofa, Chair, and Table. We also provide additional results on automatically generated synthetic scenes in the supplementary material including an additional object category.


\subsection{Evaluation Criteria}

\begin{table*}[h]
\centering
\scalebox{.86}{
\begin{tabular}{ccc@{$\quad$}c@{$\quad$}cccc}
\toprule
&Parameter& Model from~\cite{pearl2022geocode} & Model from~\cite{pearl2022geocode} & CD & CD & Genetic & Genetic \\
& & w\textbackslash o refinement & & w\textbackslash o refinement & & w\textbackslash o refinement & \\
\midrule
\parbox[t]{4mm}{\multirow{12}{*}{\rotatebox[origin=c]{85}{Continuous Parameters}}} 
& & \multicolumn{6}{c}{Mean Absolute Difference to Ground Truth for Table~(\textdownarrow)} \\
\cmidrule{3-8}
& Width [cm] & 0.74 & 0.22 & 0.21 & 0.21 & 0.15 & \textbf{0.08} \\
& Height [cm] & 0.34 & 0.14 & 0.15 & 0.15 & 0.12 & \textbf{0.06} \\
& Depth [cm] & 0.4 & 0.2 & 0.22 & 0.22 & 0.13 & \textbf{0.08} \\
& Mid Leg X Scale & 0.19 & 0.13 & 0.16 & 0.14 & 0.1 & \textbf{0.07} \\
& Mid Leg Y Scale & 0.16 & 0.14 & 0.11 & 0.11 & \textbf{0.08} & \textbf{0.08} \\
& Legs Offset Y & 0.35 & 0.34 & 0.37 & 0.37 & 0.31 & \textbf{0.27} \\
& Legs Offset X & 0.32 & 0.32 & 0.36 & 0.36 & 0.15 & \textbf{0.13} \\
& Legs Scale Y & 0.17 & 0.15 & 0.18 & 0.19 & 0.13 & \textbf{0.1} \\
& Mid Board Z Scale & 0.21 & 0.18 & 0.18 & 0.18 & 0.17 & \textbf{0.14} \\
& Cabinet Width Scale & 0.15 & 0.1 & 0.15 & 0.11 & 0.1 & \textbf{0.05} \\
& Top Thickness [cm] & 0.03 & 0.03 & \textbf{0.02} & \textbf{0.02} & \textbf{0.02} & \textbf{0.02} \\
& Legs Size [cm] & 0.03 & 0.03 & 0.03 & 0.03 & \textbf{0.01} & \textbf{0.01} \\
\midrule
\parbox[t]{4mm}{\multirow{5}{*}{\rotatebox[origin=c]{85}{Discrete P.}}}
& & \multicolumn{6}{c}{Classification Accuracy for Table~(\textuparrow)} \\
\cmidrule{3-8}
& Top Shape &0.78 & 0.78 & 0.72 & 0.72 & \textbf{0.94} & \textbf{0.94} \\
& Legs Type &0.39 & 0.39 & 0.6 & 0.58 & \textbf{0.78} & \textbf{0.78} \\
& Has Cabinet Leg &0.71 & 0.71 & 0.57 & 0.55 & 0.78 & \textbf{0.86} \\
& Has Mid Board &0.71 & 0.71 & 0.55 & 0.49 & \textbf{0.76} & \textbf{0.76} \\
\midrule
\toprule
& & \multicolumn{6}{c}{Reconstruction quality across all categories~(\textdownarrow)} \\
\cmidrule{3-8}
& Chamfer distance [m] & 0.52 & 0.21 & 0.16 & 0.16 & 0.14 & \textbf{0.11} \\
& Abs. rotation error [rad] & 0.45 & 0.12 & 0.13 & 0.11 & 0.08 & \textbf{0.06} \\
\bottomrule
\end{tabular}
}
\caption{Quantitative results of shape program parameters between different baselines for category Table, and general reconstruction metrics across all categories. We observe that GeoCode~\cite{pearl2022geocode} performs worst due to the domain gap, but the optimization helps improve the performance. Coordinate Descent~(CD) does better but is prone to local minima. Genetic algorithm outperforms other baselines by a significant margin.}
\label{tab:eval_scannet}
\vspace{-0.1cm}
\end{table*}


\textbf{Evaluation of shape parameters.}
Each object category has properties which are an important aspect of the reconstruction. This is neglected by current state-of-the-art methods. Our approach integrates these properties as shape parameters into the reconstruction and we evaluate them by comparing the reconstructed and ground truth parameters of the target objects. For the experiments with ScanNet, we manually annotated all relevant objects with shape parameters. 

\textbf{Data pre-processing.} We assume that for the target object in the RGB-D scan, we have an initial estimate for the 3D center of the target object, here we follow preprocessing from~\cite{ainetter2023hocsearch}. Then, we estimate 2D segmentation masks of the object used to generate dense point clouds in the objective function in Equation~\eqref{eq:loss}.

\subsection{Search Baseline Details}

In our evaluations of our search algorithm, we compare results of our genetic algorithm to two baselines:

\begin{itemize}
    \item GeoCode~\cite{pearl2022geocode} introduces a supervised learning method based on an encoder-decoder architecture that takes a target point cloud as input and outputs a value for target program parameters. We use DGCNN~\cite{phan2018dgcnn} as encoder, and the resulting embedding of size $128$ is shared between separate decoders that predict values for program parameters. More precisely, we use three feed-forward layers per parameter, each of size $256$. For discrete parameters, we add a classification head as the output layer with number of outputs matching the number of different discrete values. For continuous parameters, we simply regress the output value. Finally, we perform gradient descent that further refines predictions by passing predicted parameter values to PyTorchGeoNodes and optimizing the objective function. We train it by randomly generating $10000$ instances of ground truth shape program parameters and rotations and then generating synthetic point clouds from these parameters using PyTorchGeoNodes.  
    \item Our implementation of coordinate descent iteratively optimizes individual parameters of the shape program. First, we initialize the estimate of parameters randomly. Then, at every iteration, we take the current estimate and replace value of the selected parameter by all valid discrete values and calculate the objective term. In case of continuous values we first discretize them as we explain in the supplementary. In case this improves the objective, we update the current estimate. Finally, we perform gradient descent that further refines the current estimate by passing it to PyTorchGeoNodes and optimizing the objective function. We repeat the process for all different parameters and loop 3 times which approximately matches the execution time of our genetic algorithm.   
\end{itemize}

\subsection{Validation on the ScanNet Dataset}
\label{sec:eval_scannet}

\begin{figure}
    \centering
    \begin{tabular}{c@{}c@{}c}
    \includegraphics[width=0.32\linewidth,trim=11.3cm 0 0 0,clip]{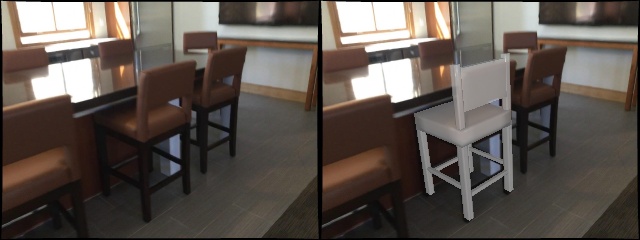} & 
    \includegraphics[width=0.32\linewidth,trim=11.3cm 0 0 0,clip]{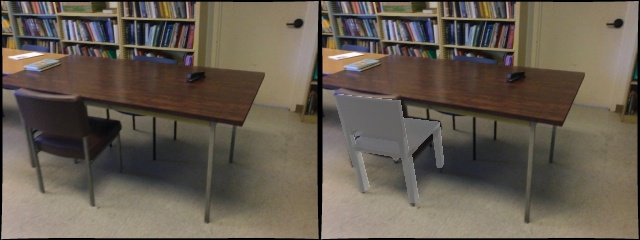} & 
    \includegraphics[width=0.32\linewidth,trim=11.3cm 0 0 0,clip]{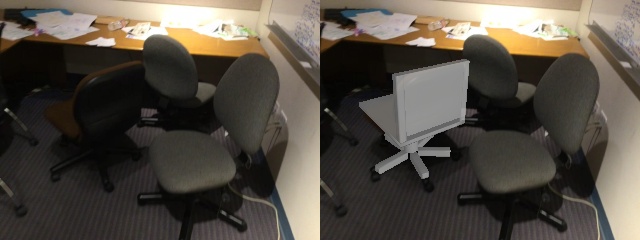} \\ 
    \includegraphics[width=0.32\linewidth,trim=11.3cm 0 0 0,clip]{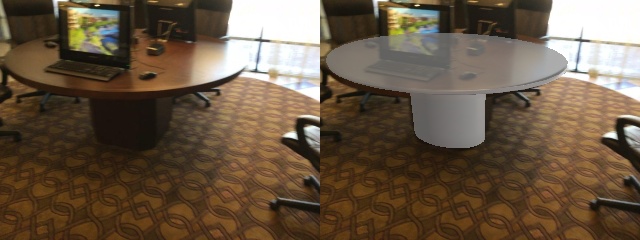} & 
    \includegraphics[width=0.32\linewidth,trim=11.3cm 0 0 0,clip]{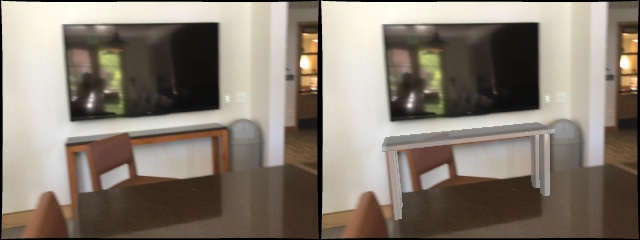} & 
    \includegraphics[width=0.32\linewidth,trim=11.3cm 0 0 0,clip]{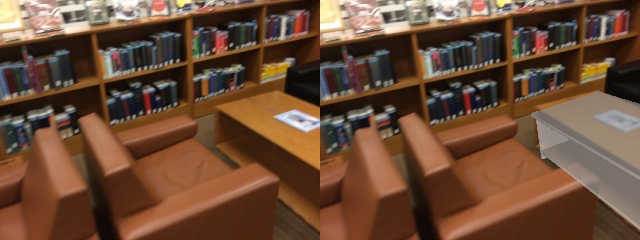} \\
    \includegraphics[width=0.32\linewidth,trim=11.3cm 0 0 0,clip]{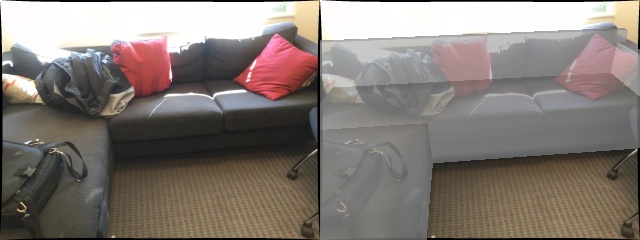} & 
    \includegraphics[width=0.32\linewidth,trim=11.3cm 0 0 0,clip]{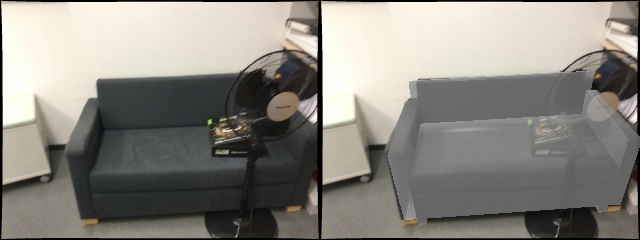} & 
    \includegraphics[width=0.32\linewidth,trim=11.3cm 0 0 0,clip]{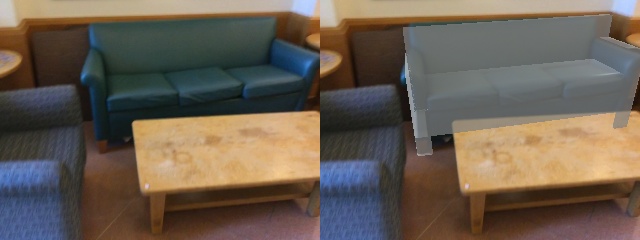} \\ 

    \end{tabular}
    \caption{Shape from recovered object parameters aligns very well with the target object also when projected into sampled images of the scene. In the first row, we reconstruct a variety of chairs along with many detailed parameters including seat thickness, thickness and offsets of side and front/back support legs, existence of different semantic elements like 'star-shaped' legs and its exact rotation. In the second row, we correctly estimate shape of table top and its measurements, including width, depth and thickness, parameters of table legs, and existence of middle shelf including its offset from the bottom. In the third row we reconstruct parameters of 'L-element' of sofa including its depth and width, existence and size of legs, and existence of individual armrests, backrest, and the corresponding measurements. We show more results in the supplementary material.}
    \label{fig:qualfig1}
\end{figure}


We evaluate our approach on 
challenging real-world scenarios on  validation scenes from ScanNet~\cite{dai2017scannet}, which provides RGB-D scans of indoor scenes. We selected a subset of object instances from the Scannotate dataset~\cite{ainetter2023hocsearch} for which we manually annotated 57 sofas, 52 chairs, and 67 tables. 



Table~\ref{tab:eval_scannet} evaluates the quality for object parameters reconstruction for the 'table' category and general reconstruction quality across different categories. We report the numbers for 'chair' and 'sofa' in the supplementary material. We measure the classification accuracy for discrete parameteres and absolute difference for continuous parameters.

 We observe that the supervised learning method from~\cite{pearl2022geocode}, trained on synthetic data, generalises poorly to the scene from ScanNet, while coordinate descent baseline is prone to local minima. Our genetic algorithm performs very well across all metrics except for parameters defining very small details. Visually, the models recovered by our method are comparable in quality to the ground truth and we confirm the benefits of our refinement procedure. Figure~\ref{fig:qualfig1} and Figure~\ref{fig:qualfig2} show some qualitative results. In the supplementary we show additional results, and include a downstream application for \textit{point cloud segmentation} based on our reconstructions.

\newlength{\niceresultheight}
\setlength{\niceresultheight}{1.5cm}
\newlength{\niceresultwidth}
\setlength{\niceresultwidth}{3.5cm}

\newcommand{\niceresult}[2]{
\multicolumn{2}{c}{\includegraphics[height=\niceresultheight]{figures/qualitative_results/#1_obj_#2/overlay_new.png}} &
\includegraphics[height=\niceresultheight]{figures/qualitative_results/#1_obj_#2/our_gaussians.png}
}

\begin{figure}
    \centering
    \scalebox{1.0}{
    \begin{tabular}{ccc}
        \niceresult{scene0187_01}{4} \\
        \niceresult{scene0474_01}{14} \\
        \niceresult{scene0277_00}{4} \\
        
        
        \hspace{0.7cm} (a) & (b) & (c) \\
    \end{tabular}}
    \vspace{-0.3cm}
    \caption{(a) We overlay shape reconstructed using our retrieved shape parameters with the image and (b) compare it to the ground truth. (c) Our integration of Gaussians into PyTorchGeoNodes enables appealing 3D reconstructions.
    }
    \label{fig:qualfig2}
    \vspace{-0.3cm}
\end{figure}

 \section{Conclusion}

Our code is publicly available and we hope this will encourage other researchers to develop the field and further extend possibilities with PyTorchGeoNodes. While procedural models / shape programs are an attractive representation of objects, there are indeed still many things to develop: \eg extensions to consider deformable and articulated objects and to integrate textures, or extensions for scene-level procedural models / programs (learned or designed in Blender for example) that would enable reasoning about object relationships and scene physics. Modularity of PyTorchGeoNodes enables integrations with other types of 3D representation, as we demonstrate with our integration of Gaussian Splatting. Moreover, differentiable rendering of PyTorchGeoNodes enables development of deep learning methods for creating shape programs from images in a self-supervised way. Finally, a major limitation is that our shape programs need to be designed by humans for different categories which can lead to limited expressiveness when applying our approach in the wild. Learning shape program designs from database of shapes would make our method easily applicable to novel categories.   


\newpage

\section*{Acknowledgement}

This project was funded by the European Union (ERC Advanced Grant explorer Funding ID \#101097259). This work was performed using HPC resources from GENCI-IDRIS (Grant 2024-AD010615585).
}

{
    \small
    \bibliographystyle{ieeenat_fullname}
    \bibliography{main}

\begin{thebibliography}{40}
\providecommand{\natexlab}[1]{#1}
\providecommand{\url}[1]{\texttt{#1}}
\expandafter\ifx\csname urlstyle\endcsname\relax
  \providecommand{\doi}[1]{doi: #1}\else
  \providecommand{\doi}{doi: \begingroup \urlstyle{rm}\Url}\fi

\bibitem[Ainetter et~al.(2024)Ainetter, Stekovic, Fraundorfer, and Lepetit]{ainetter2023hocsearch}
Stefan Ainetter, Sinisa Stekovic, Friedrich Fraundorfer, and Vincent Lepetit.
\newblock {HOC-Search: Efficient CAD Model and Pose Retrieval from RGB-D Scans}.
\newblock \emph{International Conference on 3D Vision}, 2024.

\bibitem[Aubry and Russell(2015)]{aubry2015understanding}
Mathieu Aubry and Bryan~C Russell.
\newblock {Understanding Deep Features with Computer-Generated Imagery}.
\newblock In \emph{International Conference on Computer Vision}, 2015.

\bibitem[Aubry et~al.(2014)Aubry, Maturana, Efros, Russell, and Sivic]{aubry2014seeing}
Mathieu Aubry, Daniel Maturana, Alexei~A Efros, Bryan~C Russell, and Josef Sivic.
\newblock {Seeing 3D Chairs: Exemplar Part-Based 2D-3D Alignment Using a Large Dataset of CAD Models}.
\newblock In \emph{Conference on Computer Vision and Pattern Recognition}, 2014.

\bibitem[Botsch and Kobbelt(2004)]{botsch2004intuitive}
Mario Botsch and Leif Kobbelt.
\newblock {An Intuitive Framework for Real-Time Freeform Modeling}.
\newblock \emph{ACM Trans. on Graphics~(TOG)}, 2004.

\bibitem[Cao and de~Charette(2023)]{cao2023scenerf}
Anh-Quan Cao and Raoul de Charette.
\newblock {SceneRF: Self-Supervised Monocular 3D Scene Reconstruction with Radiance Fields}.
\newblock In \emph{International Conference on Computer Vision}, 2023.

\bibitem[Chang et~al.(2015)Chang, Funkhouser, Guibas, Hanrahan, Huang, Li, Savarese, Savva, Song, Su, Xiao, Yi, and Yu]{chang2015shapenet}
Angel~X. Chang, Thomas Funkhouser, Leonidas Guibas, Pat Hanrahan, Qixing Huang, Zimo Li, Silvio Savarese, Manolis Savva, Shuran Song, Hao Su, Jianxiong Xiao, Li Yi, and Fisher Yu.
\newblock {ShapeNet: An Information-Rich 3D Model Repository}.
\newblock \emph{arXiv Preprint}, 2015.

\bibitem[Chung et~al.(2024)Chung, Oh, and Lee]{chung2024depth}
Jaeyoung Chung, Jeongtaek Oh, and Kyoung~Mu Lee.
\newblock {Depth-Regularized Optimization for 3D Gaussian Splatting in Few-Shot Images}.
\newblock In \emph{Conference on Computer Vision and Pattern Recognition}, 2024.

\bibitem[Community(2018)]{blender2018}
Blender~Online Community.
\newblock \emph{Blender - a 3D modelling and rendering package}.
\newblock Blender Foundation, Stichting Blender Foundation, Amsterdam, 2018.

\bibitem[Dai et~al.(2017)Dai, Chang, Savva, Halber, Funkhouser, and Nie{\ss}ner]{dai2017scannet}
Angela Dai, Angel~X. Chang, Manolis Savva, Maciej Halber, Thomas Funkhouser, and Matthias Nie{\ss}ner.
\newblock {ScanNet: Richly-Annotated 3D Reconstructions of Indoor Scenes}.
\newblock In \emph{Conference on Computer Vision and Pattern Recognition}, 2017.

\bibitem[Ebert et~al.(2003)Ebert, Musgrave, Peachey, Perlin, and Worley]{ebert2003texturing}
David~S Ebert, F~Kenton Musgrave, Darwyn Peachey, Ken Perlin, and Steven Worley.
\newblock \emph{{Texturing \& Modeling: a Procedural Approach}}.
\newblock Morgan Kaufmann, 2003.

\bibitem[Gkioxari et~al.(2019)Gkioxari, Malik, and Johnson]{gkioxari2019mesh}
Georgia Gkioxari, Jitendra Malik, and Justin Johnson.
\newblock {Mesh R-CNN}.
\newblock In \emph{International Conference on Computer Vision}, 2019.

\bibitem[Gu{\'e}don and Lepetit(2024)]{guedon2024sugar}
Antoine Gu{\'e}don and Vincent Lepetit.
\newblock {SuGaR: Surface-Aligned Gaussian Splatting for Efficient 3D Mesh Reconstruction and High-Quality Mesh Rendering}.
\newblock In \emph{Proceedings of the IEEE/CVF Conference on Computer Vision and Pattern Recognition}, pages 5354--5363, 2024.

\bibitem[Izadinia et~al.(2017)Izadinia, Shan, and Seitz]{izadinia2017im2cad}
Hamid Izadinia, Qi Shan, and Steven~M. Seitz.
\newblock {IM2CAD}.
\newblock In \emph{Conference on Computer Vision and Pattern Recognition}, 2017.

\bibitem[Jacobson et~al.(2010)Jacobson, Tosun, Sorkine, and Zorin]{jacobson2010mixed}
Alec Jacobson, Elif Tosun, Olga Sorkine, and Denis Zorin.
\newblock {Mixed Finite Elements for Variational Surface Modeling}.
\newblock In \emph{{Computer Graphics Forum}}, 2010.

\bibitem[Jones et~al.(2020)Jones, Barton, Xu, Wang, Jiang, Guerrero, Mitra, and Ritchie]{jones2020shapeassembly}
R~Kenny Jones, Theresa Barton, Xianghao Xu, Kai Wang, Ellen Jiang, Paul Guerrero, Niloy~J Mitra, and Daniel Ritchie.
\newblock {ShapeAssembly: Learning to Generate Programs for 3D Shape Structure Synthesis}.
\newblock \emph{ACM Transactions on Graphics (TOG)}, 2020.

\bibitem[Jones et~al.(2021)Jones, Charatan, Guerrero, Mitra, and Ritchie]{jones2021shapemod}
R~Kenny Jones, David Charatan, Paul Guerrero, Niloy~J Mitra, and Daniel Ritchie.
\newblock {ShapeMOD: Macro Operation Discovery for 3D Shape Programs}.
\newblock \emph{ACM Transactions on Graphics (TOG)}, 2021.

\bibitem[Kerbl et~al.()Kerbl, Kopanas, Leimk{\"u}hler, and Drettakis]{kerbl3Dgaussians}
Bernhard Kerbl, Georgios Kopanas, Thomas Leimk{\"u}hler, and George Drettakis.
\newblock 3d gaussian splatting for real-time radiance field rendering.
\newblock \emph{ACM Transactions on Graphics}.

\bibitem[Kirillov et~al.(2023)Kirillov, Mintun, Ravi, Mao, Rolland, Gustafson, Xiao, Whitehead, Berg, Lo, et~al.]{kirillov2023segment}
Alexander Kirillov, Eric Mintun, Nikhila Ravi, Hanzi Mao, Chloe Rolland, Laura Gustafson, Tete Xiao, Spencer Whitehead, Alexander~C Berg, Wan-Yen Lo, et~al.
\newblock {Segment Anything}.
\newblock In \emph{International Conference on Computer Vision}, 2023.

\bibitem[Koehl and Roussel(2015)]{koehl-isprs15}
M. Koehl and F. Roussel.
\newblock {Procedural Modelling for Reconstruction of Historic Monuments}.
\newblock \emph{{Annals of the Photogrammetry Remote Sensing and Spatial Information Sciences}}, 2015.

\bibitem[Kuo et~al.(2020)Kuo, Angelova, Lin, and Dai]{kuo2020mask2cad}
Weicheng Kuo, Anelia Angelova, Tsung-Yi Lin, and Angela Dai.
\newblock {Mask2CAD: 3D Shape Prediction by Learning to Segment and Retrieve}.
\newblock In \emph{European Conference on Computer Vision}, 2020.

\bibitem[Long et~al.(2023)Long, Guo, Lin, Liu, Dou, Liu, Ma, Zhang, Habermann, Theobalt, et~al.]{long2023wonder3d}
Xiaoxiao Long, Yuan-Chen Guo, Cheng Lin, Yuan Liu, Zhiyang Dou, Lingjie Liu, Yuexin Ma, Song-Hai Zhang, Marc Habermann, Christian Theobalt, et~al.
\newblock {Wonder3D: Single Image to 3D using Cross-Domain Diffusion}.
\newblock \emph{arXiv Preprint}, 2023.

\bibitem[Massa et~al.(2016)Massa, Russell, and Aubry]{massa2016deep}
Francisco Massa, Bryan~C Russell, and Mathieu Aubry.
\newblock {Deep Exemplar 2D-3D Detection by Adapting from Real to Rendered Views}.
\newblock In \emph{Conference on Computer Vision and Pattern Recognition}, 2016.

\bibitem[Mottaghi et~al.(2015)Mottaghi, Xiang, and Savarese]{mottaghi2015coarse}
Roozbeh Mottaghi, Yu Xiang, and Silvio Savarese.
\newblock {A Coarse-to-Fine Model for 3D Pose Estimation and Sub-Category Recognition}.
\newblock In \emph{Conference on Computer Vision and Pattern Recognition}, 2015.

\bibitem[M{\"u}ller et~al.(2006)M{\"u}ller, Wonka, Haegler, Ulmer, and Van~Gool]{muller2006procedural}
Pascal M{\"u}ller, Peter Wonka, Simon Haegler, Andreas Ulmer, and Luc Van~Gool.
\newblock {Procedural Modeling of Buildings}.
\newblock In \emph{ACM SIGGRAPH 2006 Papers}, 2006.

\bibitem[Park et~al.(2019)Park, Florence, Straub, Newcombe, and Lovegrove]{park-cvpr19-deepsdf}
Jeong~Joon Park, Peter Florence, Julian Straub, Richard~A. Newcombe, and Steven~J. Lovegrove.
\newblock {{DeepSDF}: Learning Continuous Signed Distance Functions for Shape Representation}.
\newblock In \emph{Conference on Computer Vision and Pattern Recognition}, 2019.

\bibitem[Paszke et~al.(2019)Paszke, Gross, Massa, Lerer, Bradbury, Chanan, Killeen, Lin, Gimelshein, Antiga, et~al.]{paszke2019pytorch}
Adam Paszke, Sam Gross, Francisco Massa, Adam Lerer, James Bradbury, Gregory Chanan, Trevor Killeen, Zeming Lin, Natalia Gimelshein, Luca Antiga, et~al.
\newblock Pytorch: An imperative style, high-performance deep learning library.
\newblock In \emph{Advances in Neural Information Processing Systems}, 2019.

\bibitem[Pearl et~al.(2022)Pearl, Lang, Hu, Yeh, and Hanocka]{pearl2022geocode}
Ofek Pearl, Itai Lang, Yuhua Hu, Raymond~A Yeh, and Rana Hanocka.
\newblock {GeoCode: Interpretable Shape Programs}.
\newblock \emph{arXiv Preprint}, 2022.

\bibitem[Phan et~al.(2018)Phan, Le~Nguyen, Nguyen, and Bui]{phan2018dgcnn}
Anh~Viet Phan, Minh Le~Nguyen, Yen Lam~Hoang Nguyen, and Lam~Thu Bui.
\newblock {DGCNN: A Convolutional Neural Network over Large-Scale Labeled Graphs}.
\newblock \emph{Neural Networks}, 2018.

\bibitem[Ravi et~al.(2020)Ravi, Reizenstein, Novotny, Gordon, Lo, Johnson, and Gkioxari]{ravi2020pytorch3d}
Nikhila Ravi, Jeremy Reizenstein, David Novotny, Taylor Gordon, Wan-Yen Lo, Justin Johnson, and Georgia Gkioxari.
\newblock Accelerating 3d deep learning with pytorch3d.
\newblock \emph{arXiv Preprint}, 2020.

\bibitem[Ritchie et~al.(2023)Ritchie, Guerrero, Jones, Mitra, Schulz, Willis, and Wu]{ritchie2023neurosymbolic}
Daniel Ritchie, Paul Guerrero, R~Kenny Jones, Niloy~J Mitra, Adriana Schulz, Karl~DD Willis, and Jiajun Wu.
\newblock {Neurosymbolic Models for Computer Graphics}.
\newblock In \emph{Computer graphics forum}. Wiley Online Library, 2023.

\bibitem[Shin et~al.(2018)Shin, Fowlkes, and Hoiem]{shin-cvpr18-pixelsvoxelsandviews}
D. Shin, Charless~C. Fowlkes, and Derek Hoiem.
\newblock {Pixels, Voxels, and Views: A Study of Shape Representations for Single View 3D Object Shape Prediction}.
\newblock In \emph{Conference on Computer Vision and Pattern Recognition}, 2018.

\bibitem[Simon et~al.(2011)Simon, Teboul, Koutsourakis, and Paragios]{simon-ijcv2011-shape}
Loic Simon, Olivier Teboul, Panagiotis Koutsourakis, and Nikos Paragios.
\newblock {Random Exploration of the Procedural Space for Single-View 3D Modeling of Buildings}.
\newblock \emph{International Journal of Computer Vision}, 2011.

\bibitem[Sorkine and Alexa(2007)]{sorkine2007rigid}
Olga Sorkine and Marc Alexa.
\newblock {As-Rigid-as-Possible Surface Modeling}.
\newblock In \emph{{Symposium on Geometry Processing}}, 2007.

\bibitem[Szymanowicz et~al.(2023)Szymanowicz, Rupprecht, and Vedaldi]{splatter}
Stanislaw Szymanowicz, Christian Rupprecht, and Andrea Vedaldi.
\newblock {Splatter Image: Ultra-Fast Single-View 3D Reconstruction}.
\newblock \emph{arXiv Preprint}, 2023.

\bibitem[Tian et~al.(2019)Tian, Luo, Sun, Ellis, Freeman, Tenenbaum, and Wu]{tian2019learning}
Yonglong Tian, Andrew Luo, Xingyuan Sun, Kevin Ellis, William~T Freeman, Joshua~B Tenenbaum, and Jiajun Wu.
\newblock {Learning to Infer and Execute 3D Shape Programs}.
\newblock \emph{International Conference for Learning Representations}, 2019.

\bibitem[Turkulainen et~al.(2024)Turkulainen, Ren, Melekhov, Seiskari, Rahtu, and Kannala]{turkulainen2024dn}
Matias Turkulainen, Xuqian Ren, Iaroslav Melekhov, Otto Seiskari, Esa Rahtu, and Juho Kannala.
\newblock {DN-Splatter: Depth and Normal Priors for Gaussian Splatting and Meshing}.
\newblock \emph{arXiv preprint arXiv:2403.17822}, 2024.

\bibitem[Vaienti et~al.(2023)Vaienti, Petitpierre, di~Lenardo, and Kaplan]{vaienti2023}
Beatrice Vaienti, Rémi Petitpierre, Isabella di Lenardo, and Frédéric Kaplan.
\newblock {Machine-Learning-Enhanced Procedural Modeling for 4D Historical Cities Reconstruction}.
\newblock \emph{Remote Sensing}, 15\penalty0 (13), 2023.

\bibitem[Wang et~al.(2018)Wang, Zhang, Li, Fu, Liu, and Jiang]{wang2018pixel2mesh}
Nanyang Wang, Yinda Zhang, Zhuwen Li, Yanwei Fu, Wei Liu, and Yu-Gang Jiang.
\newblock {Pixel2Mesh: Generating 3D Mesh Models from Single RGB Images}.
\newblock In \emph{European Conference on Computer Vision}, 2018.

\bibitem[Wonka et~al.(2003)Wonka, Wimmer, Sillion, and Ribarsky]{wonka2003instant}
Peter Wonka, Michael Wimmer, Fran{\c{c}}ois Sillion, and William Ribarsky.
\newblock {Instant Architecture}.
\newblock \emph{ACM Transactions on Graphics (TOG)}, 2003.

\bibitem[Wu et~al.(2024)Wu, Yuan, Zhang, Yang, Cao, Yan, and Gao]{wu2024recent}
Tong Wu, Yu-Jie Yuan, Ling-Xiao Zhang, Jie Yang, Yan-Pei Cao, Ling-Qi Yan, and Lin Gao.
\newblock {Recent Advances in 3D Gaussian Splatting}.
\newblock \emph{Computational Visual Media}, 2024.

\end{thebibliography}


\begin{thebibliography}{4}
\providecommand{\natexlab}[1]{#1}
\providecommand{\url}[1]{\texttt{#1}}
\expandafter\ifx\csname urlstyle\endcsname\relax
  \providecommand{\doi}[1]{doi: #1}\else
  \providecommand{\doi}{doi: \begingroup \urlstyle{rm}\Url}\fi

\bibitem[Ainetter et~al.(2024)Ainetter, Stekovic, Fraundorfer, and Lepetit]{ainetter2023hocsearch}
Stefan Ainetter, Sinisa Stekovic, Friedrich Fraundorfer, and Vincent Lepetit.
\newblock {HOC-Search: Efficient CAD Model and Pose Retrieval from RGB-D Scans}.
\newblock \emph{International Conference on 3D Vision}, 2024.

\bibitem[Dai et~al.(2017)Dai, Chang, Savva, Halber, Funkhouser, and Nie{\ss}ner]{dai2017scannet}
Angela Dai, Angel~X. Chang, Manolis Savva, Maciej Halber, Thomas Funkhouser, and Matthias Nie{\ss}ner.
\newblock {ScanNet: Richly-Annotated 3D Reconstructions of Indoor Scenes}.
\newblock In \emph{Conference on Computer Vision and Pattern Recognition}, 2017.

\bibitem[Gu{\'e}don and Lepetit(2024)]{guedon2024sugar}
Antoine Gu{\'e}don and Vincent Lepetit.
\newblock {SuGaR: Surface-Aligned Gaussian Splatting for Efficient 3D Mesh Reconstruction and High-Quality Mesh Rendering}.
\newblock In \emph{Proceedings of the IEEE/CVF Conference on Computer Vision and Pattern Recognition}, pages 5354--5363, 2024.

\bibitem[Pearl et~al.(2022)Pearl, Lang, Hu, Yeh, and Hanocka]{pearl2022geocode}
Ofek Pearl, Itai Lang, Yuhua Hu, Raymond~A Yeh, and Rana Hanocka.
\newblock {GeoCode: Interpretable Shape Programs}.
\newblock \emph{arXiv Preprint}, 2022.

\end{thebibliography}
}


\end{document}


\maketitle

We provide additional results and details regarding evaluations of our search algorithm and other baselines in Section~\ref{sec:validation}, and provide more details regarding our PyTorchGeoNodes framework in Section~\ref{sec:pytorchgeonodes}, regarding our search algorithm in Section~\ref{sec:search}, regarding our integration of Gaussian splats in Section~\ref{sec:gaussian} and regarding our shape program designs in Section~\ref{sec:designs}. 

\textbf{In addition, we provide a \textbf{video} showing the reconstruction progress of our approach through genetic iterations for scenes of the ScanNet dataset~\cite{dai2017scannet}, qualitative results for our integration of Gaussian splats, and a demo on shape manipulation with PyTorchGeoNodes and Gaussian splats.}

\newcommand{\niceresult}[2]{
\multicolumn{2}{c}{\includegraphics[width=0.33\linewidth]{supplementary/#1_qual/ex_#2.jpg}}}

\section{Additional Validations}
\label{sec:validation}

In this section, we first provide more details regarding baseline implementations. Then, we provide additional evaluations of our genetic algorithm for the ScanNet dataset and our synthetic dataset. Finally, we provide additional validation for our integration of Gaussian splats into our pipeline.

\subsection{Additional Validation on ScanNet}

Tables~\ref{tab:eval_sp_sofa},~\ref{tab:eval_sp_chair} show quantitative evaluations on ScanNet. We compare 3D shapes of our recovered program parameters to 3D shapes of ground truth program parameters using chamfer distance between the corresponding point clouds.

We confirm that our proposed genetic algorithm outperforms alternative baselines. Direct inference with deep learning from~\cite{pearl2022geocode} struggles to deal with the domain gap and does not generalize to partial point clouds in the ScanNet dataset. While refinement using our PyTorchGeoNodes still helps to improve results, in most cases, the reconstructed parameters and 3D shapes are not representative of the input scene. While our implementation of coordinate descent performs better, it has similar drawbacks, and fails to provide reasonable reconstruction especially in cases of bad random initialisation. In contrast, we observed substantial improvements when relying on our genetic algorithm to recover shape parameters. We observe very good performance, except in exceptions such as in the case of sofa legs where our objective function is not well-tailored to handle such small details in the scene.

\begin{figure}
    \centering
    \begin{tabular}{cc}
         \includegraphics[width=0.9\linewidth]{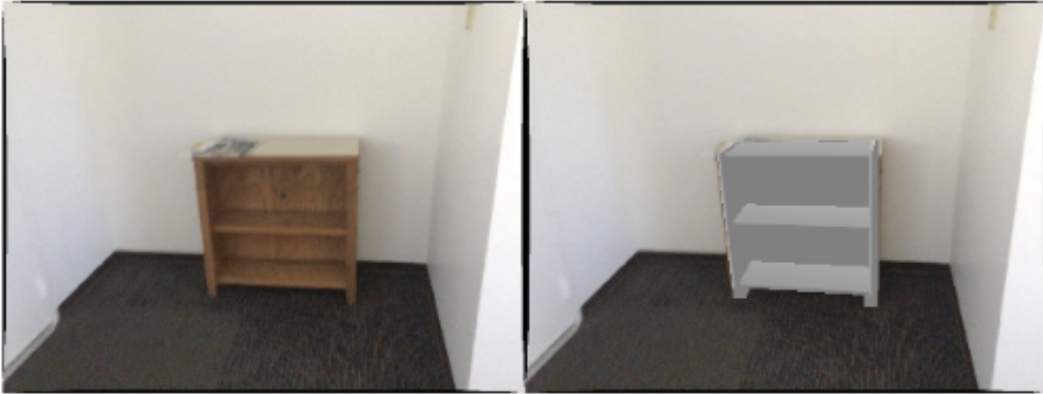}
    \end{tabular}
    \caption{Qualitative result on cabinet. Our search algorithm is able to recover correct parameters for 'Cabinet' program  that includes different number of shelves, existence of doors, size of cabinet, and legs parameters.}
    \label{fig:qual-cabinet}
\end{figure}

Qualitative results in Figures~\ref{fig:qual-chairs},~\ref{fig:qual-sofa},~\ref{fig:qual-table} show additional examples where our approach is able to reconstruct a variety of program parameters that are geometrically consistent with the input scene. In addition, we provide qualitative result for 'Cabinet' class in Figure~\ref{fig:qual-cabinet} to show our approach can be extended to other categories. We note that we were not able to provide meaningful quantitative results for this category on ScanNet: This is because the large majority of cabinets have drawers, therefore the back of cabinets are not visible and we cannot estimate the number of dividing boards based on geometric objective terms in the presence of drawers. In the case of bookshelves, objects on shelves induce noise in instance segmentation. Therefore, the objective term does not handle such settings well. We note these are general limitations for reconstructing cabinets and are present also for state-of-the-art methods~\cite{ainetter2023hocsearch}. 

\begin{figure}
    \centering
    \scalebox{0.9}{
    \begin{tabular}{cccc}
         \includegraphics[width=0.17\linewidth]{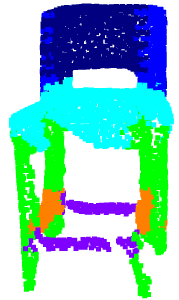} &
         \includegraphics[width=0.21\linewidth]{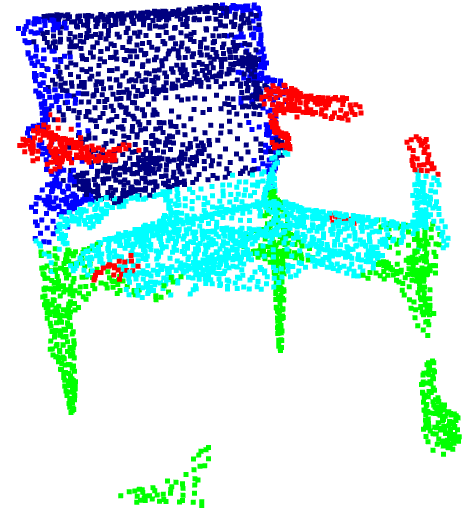} &
         \includegraphics[width=0.21\linewidth]{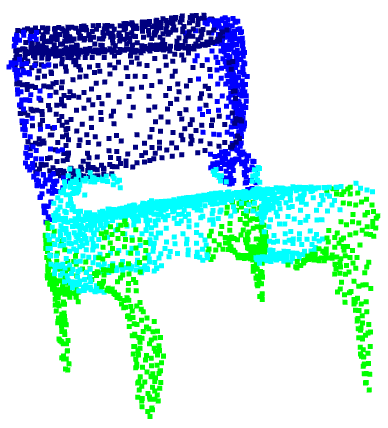} &
         \includegraphics[width=0.21\linewidth]{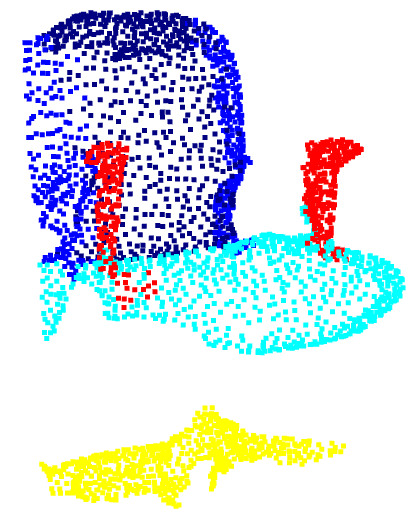} \\
         \multicolumn{2}{c}{\includegraphics[width=0.3\linewidth]{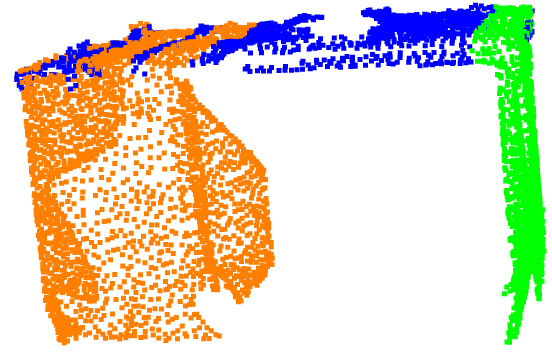}} & 
         \multicolumn{2}{c}{\includegraphics[width=0.4\linewidth,trim=0 0 0 0.5cm,clip]{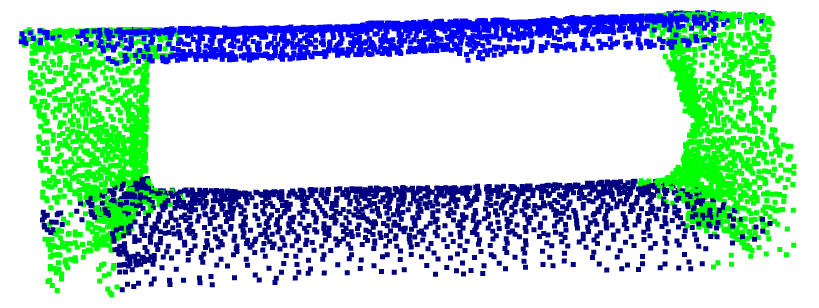}} \\
         \multicolumn{2}{c}{\includegraphics[width=0.4\linewidth]{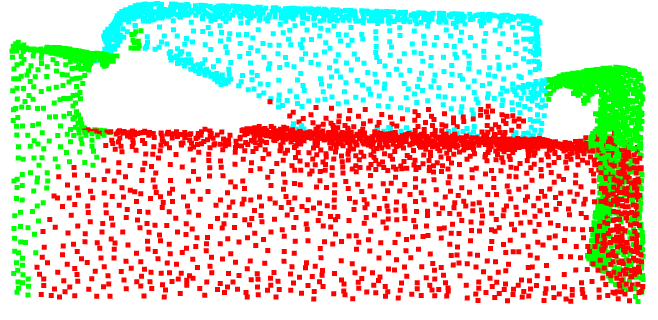}} &
         \multicolumn{2}{c}{\includegraphics[width=0.4\linewidth]{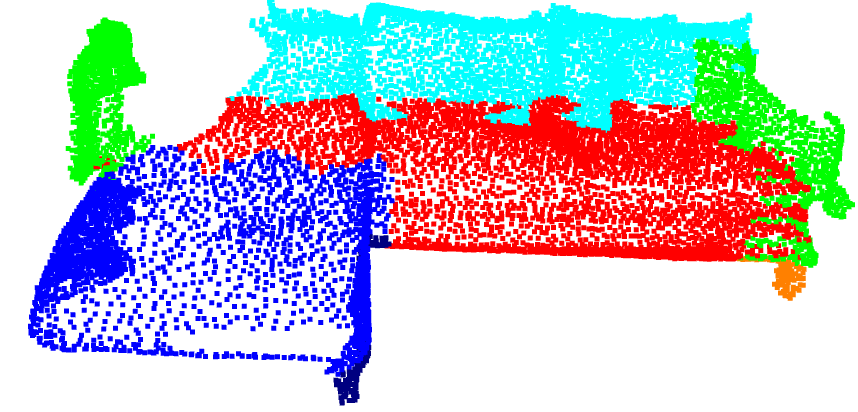}} \\
    \end{tabular}}
    \caption{Segmenting semantic parts in partial point clouds of objects based on our reconstructions.}
    \label{fig:pointcloudseg}
\end{figure}

We show in Figure~\ref{fig:pointcloudseg} that our reconstructions can also be used for segmenting semantic object parts in partial point clouds. In this application, graph of PyTorchGeoNodes directly assigns points in the point cloud to base primitives in our procedural graphs based on simple chamfer distance.

\subsection{Validation on Synthetic Data}
\label{sec:synthetic}

In addition, we evaluate our search algorithm on synthetic scenes to evaluate performance of our algorithm when there is no presence of occlusions and noise in the data. We generate $300$ synthetic scenes that contain complete point clouds of objects corresponding to target parameters. In this experiment, we simply use chamfer distance loss as we do not need to handle occlusions in the scene. We show quantitative results in Tables~\ref{tab:synth_cabinet},~\ref{tab:synth_chair} for 'Cabinet' and 'Chair'. We omit 'Sofa', and 'Table' categories from the supplementary where we observed similar behavior. Similarly to our experiments on ScanNet, we confirm that our genetic algorithm performs extremely well, and we demonstrate advantage of adding gradient descent based on our PyTorchGeoNodes. However, we also observe significant boost in performance compared to our ScanNet experiments. This is a clear indicator that introducing novel objective terms into the search could further improve the reconstruction results.

\subsection{Validation of Gaussian Splats Integration}

\begin{figure}[ht]
    \centering
    \scalebox{0.8}{
    \begin{tabular}{cc}
         \multicolumn{2}{c}{\includegraphics[width=0.9\linewidth]{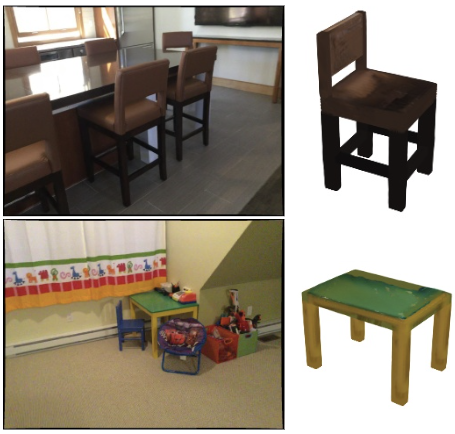}} \\
         \hspace{1.5cm} Input view & \hspace{1cm} Our results \\ 
    \end{tabular}}
    \caption{Qualitative results for our Gaussian splats integration into PyTorchGeoNodes. We show more results in the supplementary video. We point to the legs of the chair and table. In the input views, there are self-occlusions in case of chair, or severe occlusions from other object. Our computational-graph-aware is able to deal with such situations and we reconstruct appearance of these parts of objects accurately.}
    \label{fig:qual-gauss}
\end{figure}

\begin{figure}[ht]
    \centering
    \begin{tabular}{cc}
         \includegraphics[width=0.4\linewidth,trim=5.5cm 4.5cm 5cm 6cm,clip]{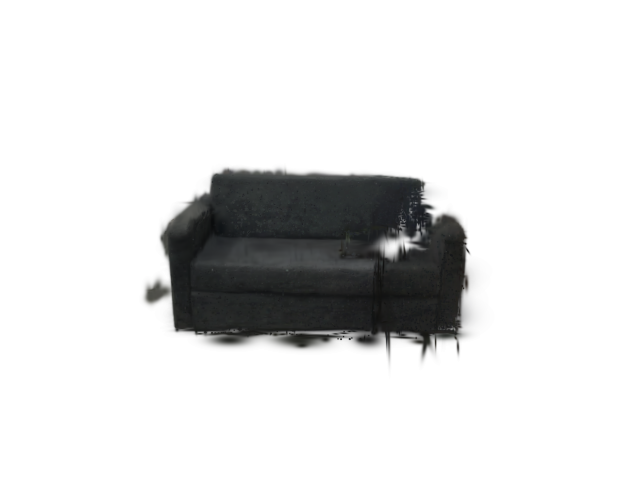} & \includegraphics[width=0.4\linewidth,trim=5.5cm 4.5cm 5cm 6cm,clip]{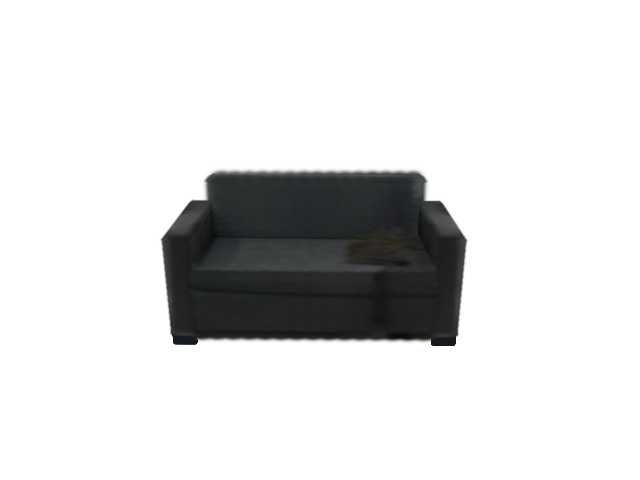}\\ 
         Vanilla Gaussian splatting & Our procedural Gaussians 
    \end{tabular}
    \caption{Vanilla Gaussian splatting on target objects does not perform well because of occlusions, lack of views of the target object, and noisy observations. In contrast, our integration of Gaussian splatting and PyTorchGeoNodes results in reconstructions of much higher quality, as Gaussians are constrained by the corresponding procedural model.}
    \label{fig:qual-gauss2}
\end{figure}

\begin{table}
\centering
\begin{tabular}{ccc}
\toprule
& Initialisation only & Final \\
\midrule
SSIM ($\uparrow$) & 0.96 & \textbf{0.99} \\
PSNR ($\uparrow$) & 31.1 & \textbf{36.74} \\
LPIPS ($\downarrow$) & 0.06 & \textbf{0.03} \\
\bottomrule
\end{tabular}
\caption{Quantitative results on standard appearance metrics of our Gaussian splats integration into PyTorchGeoNodes.}
\label{tab:eval_gaussian}
\end{table}

In Table~\ref{tab:eval_gaussian} we show quantitative evaluations of our Gaussian splats integration into PyTorchGeoNodes using standard metrics and we mask pixel locations that do not belong to target objects. For the quantitative evaluation, we randomly select 20 views for training and 10 for validation per secene. We selected $10$ samples among different categories for this ablation. We choose samples with masks of satisfying quality and accurate shape parameters quality for this experiment for fair evaluation as Gaussian splatting is very sensitive to noise, especially given that we only use 30 views per scene for optimization. We observe that already after initialization the metrics are relatively high and the numbers are further improved after our optimization procedure. In Figure~\ref{fig:qual-gauss} we show qualitative results, and in Figure~\ref{fig:qual-gauss2} we compare our results to 'vanilla' Gaussian splatting that is not constrained by our procedural model. 


\begin{figure*}
    \centering
    \begin{tabular}{cccccc}
         \niceresult{sofa}{1} & \niceresult{sofa}{2} & \niceresult{sofa}{3} \\
         \niceresult{sofa}{4} & \niceresult{sofa}{5} & \niceresult{sofa}{6} \\
         \niceresult{sofa}{7} & \niceresult{sofa}{8} & \niceresult{sofa}{9} \\
         \niceresult{sofa}{10} & \niceresult{sofa}{11} & \niceresult{sofa}{12} \\
    \end{tabular}
    \caption{Qualitative results on sofas. or each pair, the left image shows one of the input views. The right images in pairs show our projections of recovered shape parameters. Our results are accurate, dimensions of sofa vary greatly in our validation set, yet we are able to reconstruct these measuremenets accurately. In addition, we accurately model existence and measurements of armrests, existence and measurements of L-extensions.}
    \label{fig:qual-sofa}
\end{figure*}

\begin{table*}
\centering
\scalebox{.7}{
    \begin{tabular}{ccc@{$\quad$}c@{$\quad$}cccc}
    \toprule
    &Parameter& GeoCode~\cite{pearl2022geocode} & GeoCode~\cite{pearl2022geocode} & CD & CD & Genetic & Genetic \\
    & & w\textbackslash o refinement & & w\textbackslash o refinement & & w\textbackslash o refinement & \\
    \midrule
    \parbox[t]{4mm}{\multirow{12}{*}{\rotatebox[origin=c]{85}{Continuous Parameters}}} 
    & & \multicolumn{6}{c}{Mean Absolute Difference to Ground Truth~(\textdownarrow)} \\
    \cmidrule{3-8}
    & Width & 0.5 & 0.27 & 0.18 & 0.15 & 0.15 & \textbf{0.12} \\
    & Height & 0.12 & 0.08 & 0.08 & 0.07 & 0.08 & \textbf{0.06} \\
    & Depth & 0.07 & \textbf{0.06} & 0.12 & 0.07 & 0.11 & \textbf{0.07} \\
    & Back Height & 0.21 & 0.12 & 0.12 & \textbf{0.05} & 0.08 & 0.06 \\
    & Back Depth & 0.07 & 0.08 & 0.07 & \textbf{0.06} & 0.07 & 0.07 \\
    & Back Over-Width Scale & 0.36 & 0.35 & 0.39 & \textbf{0.34} & 0.39 & 0.38 \\
    & L Depth & 0.17 & 0.2 & 0.13 & 0.26 & \textbf{0.11} & \textbf{0.11} \\
    & L Width & 0.07 & \textbf{0.05} & 0.09 & 0.07 & 0.09 & \textbf{0.05} \\
    & Arm Width & 0.09 & 0.08 & 0.06 & 0.05 & 0.06 & \textbf{0.04} \\
    & Arm Depth & 0.16 & 0.14 & \textbf{0.04} & 0.06 & \textbf{0.04} & 0.06 \\
    & Arm Height & 0.19 & 0.23 & 0.23 & 0.17 & 0.18 & \textbf{0.14} \\
    & Leg Size & \textbf{0.03} & 0.04 & 0.04 & 0.04 & 0.04 & 0.04 \\
    & Leg Height & 0.06 & 0.06 & 0.04 & \textbf{0.03} & 0.04 & \textbf{0.03} \\
    
    \midrule
    \parbox[t]{4mm}{\multirow{8}{*}{\rotatebox[origin=c]{85}{Discrete P.}}}
    & & \multicolumn{6}{c}{Classification Accuracy~(\textuparrow)} \\
    \cmidrule{3-8}
    & Has Back &0.4 & 0.4 & 0.77 & 0.96 & \textbf{1.0} & \textbf{1.0} \\
    & Is L-Shaped &0.56 & 0.56 & 0.77 & 0.72 & \textbf{1.0} & \textbf{1.0} \\
    & Flip L Around Y &0.6 & 0.6 & 0.9 & 0.7 & \textbf{1.0} & \textbf{1.0} \\
    & Has Left Arm &0.28 & 0.28 & 0.74 & 0.75 & 0.95 & \textbf{0.98} \\
    & Has Right Arm &0.26 & 0.26 & 0.67 & 0.75 & 0.95 & \textbf{0.98} \\
    & Has Legs &0.46 & 0.46 & 0.37 & \textbf{0.47} & 0.46 & 0.46 \\
    \bottomrule
    \end{tabular}
    }
    \caption{Quantitative results on the sofa category.}
    \label{tab:eval_sp_sofa}
\end{table*}

\begin{figure*}
    \centering
    \begin{tabular}{cccccc}
         \niceresult{chair}{1} & \niceresult{chair}{2} & \niceresult{chair}{3} \\
         \niceresult{chair}{4} & \niceresult{chair}{5} & \niceresult{chair}{6} \\
         \niceresult{chair}{7} & \niceresult{chair}{8} & \niceresult{chair}{9} \\
         \niceresult{chair}{10} & \niceresult{chair}{11} & \niceresult{chair}{12} \\
    \end{tabular}
    \caption{Qualitative results on chairs. For each pair, the left image shows one of the input views. The right images in pairs show our projections of recovered shape parameters. They are accurate, we accurately model thickness of legs, existence position and measurements of leg supports, existence, rotations and measurements of star-shape legs, and existence and measurements of armrests.}
    \label{fig:qual-chairs}
\end{figure*}

\begin{table*}
\centering
\scalebox{.7}{
\begin{tabular}{ccc@{$\quad$}c@{$\quad$}cccc}
\toprule
&Parameter& GeoCode~\cite{pearl2022geocode} & GeoCode~\cite{pearl2022geocode} & CD & CD & Genetic & Genetic \\
& & w\textbackslash o refinement & & w\textbackslash o refinement & & w\textbackslash o refinement & \\
\midrule
\parbox[t]{4mm}{\multirow{12}{*}{\rotatebox[origin=c]{85}{Continuous Parameters}}} 
& & \multicolumn{6}{c}{Mean Absolute Difference to Ground Truth~(\textdownarrow)} \\
\cmidrule{3-8}
& Seat Width & 0.24 & 0.08 & 0.06 & \textbf{0.03} & 0.06 & \textbf{0.03} \\
& Seat Height & 0.13 & 0.1 & 0.1 & 0.05 & 0.08 & \textbf{0.04} \\
& Seat Thickness & 0.05 & 0.04 & \textbf{0.03} & 0.04 & \textbf{0.03} &\textbf{ 0.03} \\
& Seat Depth & 0.14 & \textbf{0.05} & 0.07 & 0.06 & 0.06 & \textbf{0.05} \\
& Backrest Scale & 0.39 & 0.29 & 0.22 & 0.27 & 0.12 & \textbf{0.09} \\
& Back Height & 0.1 & 0.06 & 0.05 & 0.04 & 0.05 & \textbf{0.03} \\
& Back Thickness & 0.02 & 0.03 & \textbf{0.01} & 0.02 & \textbf{0.01} & \textbf{0.01} \\
& Backrest Offset Scale & 0.33 & 0.34 & 0.32 & 0.3 & 0.29 & \textbf{0.26} \\
& Legs Size & 0.03 & 0.02 & 0.03 & 0.02 & 0.03 & \textbf{0.01} \\
& Bottom Size Scale & 0.37 & 0.37 & 0.15 & 0.38 & 0.24 & \textbf{0.1} \\
& Bottom Thickness & 0.02 & 0.02 & 0.03 & \textbf{0.01} & 0.02 & 0.02 \\
& Middle Offset 2 & 0.11 & 0.12 & 0.13 & 0.14 & 0.14 & \textbf{0.05} \\
& Middle Offset 1 & 0.13 & 0.11 & 0.15 & 0.14 & 0.15 & \textbf{0.05} \\
& Middle Support Thickness & \textbf{0.01} & \textbf{0.01} & \textbf{0.01} & \textbf{0.01} & \textbf{0.01} & \textbf{0.01} \\
& Star Rotation & 0.36 & 0.36 & 0.19 & 0.42 & 0.38 & \textbf{0.1} \\
& Arm Height & 0.06 & 0.06 & 0.06 & 0.05 & 0.06 & \textbf{0.03} \\
& Arm Depth Scale & 0.11 & 0.11 & 0.1 & 0.09 & 0.1 & \textbf{0.05} \\
& Arm Width & 0.03 & 0.03 & 0.02 & \textbf{0.02} & \textbf{0.02} & 0.03 \\
& Arm Thickness & \textbf{0.01} & \textbf{0.01} & \textbf{0.01} & \textbf{0.01} & \textbf{0.01} & 0.02 \\

\midrule
\parbox[t]{4mm}{\multirow{5}{*}{\rotatebox[origin=c]{85}{Discrete P.}}}
& & \multicolumn{6}{c}{Classification Accuracy~(\textuparrow)} \\
\cmidrule{3-8}
& Has Back &0.61 & 0.6 & 0.67 & 0.67 & \textbf{1.0} & \textbf{1.0 }\\
& Legs Type &0.63 & 0.62 & 0.89 & 0.96 & \textbf{0.94} & \textbf{0.94} \\
& Has Middle Support &0.53 & 0.55 & 0.91 & 0.73 & \textbf{0.91} & \textbf{0.91} \\
& Has Arms &0.44 & 0.44 & 0.56 & 0.69 & 0.92 & \textbf{0.94} \\
\bottomrule
\end{tabular}
}
    \caption{Quantitative results on the chair category.}
    \label{tab:eval_sp_chair}
\end{table*}

\begin{figure*}
    \centering
    \begin{tabular}{cccccc}
         \niceresult{table}{1} & \niceresult{table}{2} & \niceresult{table}{3} \\
         \niceresult{table}{4} & \niceresult{table}{5} & \niceresult{table}{6} \\
         \niceresult{table}{7} & \niceresult{table}{8} & \niceresult{table}{9} \\
         \niceresult{table}{10} & \niceresult{table}{11} & \niceresult{table}{12} \\
    \end{tabular}
    \caption{Qualitative results on tables. or each pair, the left image shows one of the input views. The right images in pairs show our projections of recovered shape parameters. Our recovered shape parameters are accurate, dimensions of table vary greatly in our validation set, yet we are able to reconstruct these measurements accurately. In addition, we accurately model existence and position of middle support, existence and measurements of internal cabinet, and shape of the top board.}
    \label{fig:qual-table}
\end{figure*}



\begin{table*}
\centering
\scalebox{.7}{
    \begin{tabular}{ccc@{$\quad$}c@{$\quad$}cccc}
    \toprule
    &Parameter& GeoCode~\cite{pearl2022geocode} & GeoCode~\cite{pearl2022geocode} & CD & CD & Genetic & Genetic \\
    & & w\textbackslash o refinement & & w\textbackslash o refinement & & w\textbackslash o refinement & \\
    \midrule
    \parbox[t]{4mm}{\multirow{9}{*}{\rotatebox[origin=c]{85}{Continuous Parameters}}} 
    & & \multicolumn{6}{c}{Mean Absolute Difference to Ground Truth~(\textdownarrow)} \\
    \cmidrule{3-8}
& Height & 0.05 & \textbf{0.0} & 0.05 & 0.01 & 0.02 & \textbf{0.0} \\
& Width & 0.04 & 0.01 & 0.05 & 0.01 & 0.02 & \textbf{0.0} \\
& Depth & 0.02 & 0.02 & 0.05 & 0.0 & 0.03 & \textbf{0.0} \\
& Board Thickness & \textbf{0.0} & \textbf{0.0} & 0.02 & 0.0 & 0.02 & \textbf{0.0} \\
& Dividing Board Thickness & \textbf{0.0} & \textbf{0.0} & 0.01 & \textbf{0.0} & 0.01 & \textbf{0.0} \\
& Leg Width & 0.01 & \textbf{0.0} & 0.02 & 0.01 & 0.02 & 0.0 \\
& Leg Height & 0.01 & 0.01 & 0.02 & 0.02 & 0.02 & \textbf{0.0} \\
& Leg Depth & \textbf{0.01} & \textbf{0.01} & 0.02 & 0.02 & 0.02 & \textbf{0.01} \\
    
    \midrule
    \parbox[t]{4mm}{\multirow{10}{*}{\rotatebox[origin=c]{85}{Discrete Parameters}}}
    & & \multicolumn{6}{c}{Classification Accuracy~(\textuparrow)} \\
    \cmidrule{3-8}
& Has Drawers & \textbf{1.0} & \textbf{1.0} & 0.87 & 0.83 & 0.97 & 0.97 \\
& Number of Dividing Boards & 1.0 & 1.0 & 1.0 & 1.0 & 1.0 & 1.0 \\
& Has Back & \textbf{1.0 }& \textbf{1.0} & 0.67 & 0.67 &\textbf{ 1.0} & 0.92 \\
& Has Legs & \textbf{1.0} & \textbf{1.0} & \textbf{1.0} & 0.57 & 0.97 & \textbf{1.0} \\
& Has Drawers & \textbf{1.0} & \textbf{1.0} & 0.87 & 0.83 & 0.97 & 0.97 \\
& Number of Dividing Boards &  1.0 & 1.0 & 1.0 & 1.0 & 1.0 & 1.0 \\
& Has Back &  \textbf{1.0} &\textbf{ 1.0} & 0.67 & 0.67 & \textbf{1.0} & \textbf{1.0 }\\
& Has Legs &  \textbf{1.0} & \textbf{1.0} & \textbf{1.0} & 0.57 & 0.97 & \textbf{1.0} \\
    \bottomrule
    \end{tabular}
    }
    \caption{Quantitative results on our synthetic dataset for the cabinet category. GeoCode~\cite{pearl2022geocode} was overfitted to the data from the same distribution and therefore it reaches very high performance but it still benefits from additional refinement enabled by our PyTorchGeoNodes. Our genetic algorithm outperforms coordinate descent and performs comparably to the GeoCode baseline.}
    \label{tab:synth_cabinet}
\end{table*}

\begin{table*}
\centering
\scalebox{.7}{
    \begin{tabular}{ccc@{$\quad$}c@{$\quad$}cccc}
    \toprule
    &Parameter& GeoCode~\cite{pearl2022geocode} & GeoCode~\cite{pearl2022geocode} & CD & CD & Genetic & Genetic \\
    & & w\textbackslash o refinement & & w\textbackslash o refinement & & w\textbackslash o refinement & \\
    \midrule
    \parbox[t]{4mm}{\multirow{12}{*}{\rotatebox[origin=c]{85}{Continuous Parameters}}} 
    & & \multicolumn{6}{c}{Mean Absolute Difference to Ground Truth~(\textdownarrow)} \\
    \cmidrule{3-8}
& Seat Width & 0.04 & \textbf{0.0} & 0.06 & 0.01 & 0.05 & \textbf{0.0} \\
& Seat Height & 0.01 & \textbf{0.0} & 0.06 & 0.02 & 0.03 & \textbf{0.0} \\
& Seat Thickness & \textbf{0.0} & \textbf{0.0} & 0.04 & \textbf{0.0} & 0.04 & \textbf{0.0} \\
& Seat Depth & \textbf{0.01} & \textbf{0.01} & 0.05 & \textbf{0.01} & 0.04 & \textbf{0.01} \\
& Backrest Scale & \textbf{0.03} & 0.28 & 0.18 & 0.12 & 0.19 & \textbf{0.03} \\
& Back Height & 0.01 & 0.03 & 0.05 & 0.03 & 0.04 & \textbf{0.0} \\
& Back Thickness & \textbf{0.0} & 0.01 & 0.02 & \textbf{0.0} & 0.02 & \textbf{0.0} \\
& Backrest Offset Scale & \textbf{0.04} & 0.11 & 0.16 & 0.16 & 0.14 & 0.07 \\
& Legs Size & 0.01 & 0.01 & 0.02 & 0.0 & 0.02 & \textbf{0.0} \\
& Bottom Size Scale & \textbf{0.13} & 0.1 & 0.18 & 0.32 & 0.27 & 0.17 \\
& Bottom Thickness & \textbf{0.0} & \textbf{0.0} & 0.01 & 0.01 & 0.01 & \textbf{0.0} \\
& Middle Offset 2 & 0.06 & 0.05 & 0.08 & 0.12 & 0.1 & \textbf{0.05} \\
& Middle Offset 1 & \textbf{0.02 }& \textbf{0.02} & 0.24 & 0.18 & 0.16 & 0.06 \\
& Middle Support Thickness & \textbf{0.0} & \textbf{0.0} & 0.01 & 0.01 & 0.01 & \textbf{0.0} \\
& Star Rotation & \textbf{0.07} & 0.12 & 0.23 & 0.25 & 0.21 & 0.27 \\
& Arm Height & \textbf{0.0} & \textbf{0.0} & 0.03 & 0.01 & 0.03 & \textbf{0.0} \\
& Arm Depth Scale & 0.02 & 0.05 & 0.08 & 0.01 & 0.08 & 0.01 \\
& Arm Width & \textbf{0.0 }& 0.01 & 0.02 & \textbf{0.0} & 0.02 & \textbf{0.0} \\
& Arm Thickness & \textbf{0.0} & \textbf{0.0} & 0.01 & 0.01 & 0.01 & \textbf{0.0 }\\

    \midrule
    \parbox[t]{4mm}{\multirow{5}{*}{\rotatebox[origin=c]{85}{Discrete P.}}}
    & & \multicolumn{6}{c}{Classification Accuracy~(\textuparrow)} \\
    \cmidrule{3-8}
& Has Back &\textbf{1.0} & \textbf{1.0} & 0.9 & 0.93 & \textbf{1.0} & \textbf{1.0} \\
& Legs Type &1.0 & 1.0 & 1.0 & 1.0 & 1.0 & 1.0 \\
& Has Middle Support &\textbf{1.0} & \textbf{1.0} & 0.67 & 0.56 & 0.89 & \textbf{1.0} \\
& Has Arms &\textbf{1.0} & \textbf{1.0} & 0.93 & 0.97 & \textbf{1.0} & \textbf{1.0} \\

    \bottomrule
    \end{tabular}
    }
    \caption{Quantitative results on our synthetic dataset for the chair category. GeoCode~\cite{pearl2022geocode} was overfitted to the data from the same distribution and therefore it reaches very high performance but it still benefits from additional refinement enabled by our PyTorchGeoNodes. Our genetic algorithm outperforms coordinate descent and performs comparably to the GeoCode baseline.}
    \label{tab:synth_chair}
\end{table*}

\newpage~\newpage~\newpage~\newpage~\newpage~\newpage~\newpage~\newpage~\newpage

\section{PyTorchGeoNodes -- Implementation Details}
\label{sec:pytorchgeonodes}

In this section, we provide additional details regarding the implementation of our PyTorchGeoNodes framework. 

\subsection{Computational Graph}



Shape programs in PyTorchGeoNodes are represented as computational graphs. Therefore, for every functionality in the main paper, PyTorchGeoNodes implements a class with the corresponding functionality. 

These functionalities are implemented in the form of nodes and edges:
\begin{itemize}
    \item Nodes in our graphs are child classes of PyTorch base class \codeword{torch.nn.Module} to enable seamless integration into PyTorch code. Every node in the graph is associated with a unique id. When performing a forward pass, a node can take either default constants or outputs of other nodes as input. Therefore, nodes can contain multiple input and output sockets to enable flow of information through the computational graph.
    \item Edge is a data structure with four attributes. 'Input node' and 'Output node' define the identifiers of individual nodes that are connected by the edge. 'Input socket' and 'Output socket' define the corresponding sockets. Therefore, a node can be associated with several input and output edges, and an edge is always shared between exactly two nodes.
\end{itemize}

\begin{figure*}
    \centering
    \includegraphics[width=0.99\linewidth]{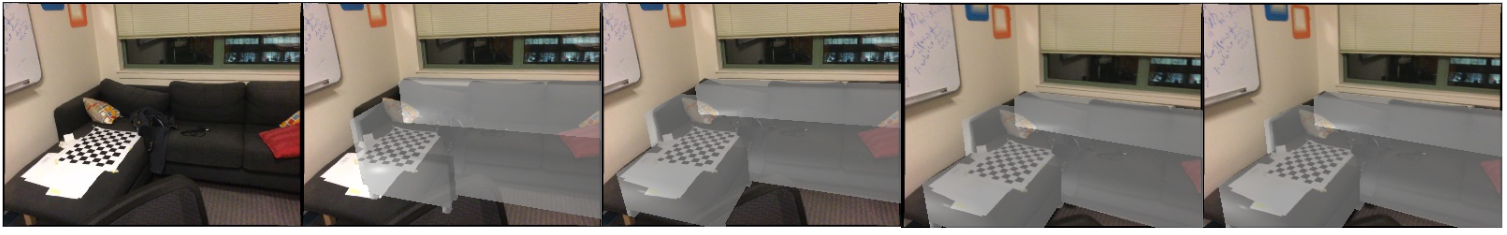}
    \caption{\textbf{Gradient-based optimization of continuous parameters of a shape program for the sofa category.} From an initial estimate of the parameters of the object, we can perform gradient descent on the parameters based on a 3D geometric loss term. In contrast to methods that directly optimize the reconstructed mesh, PyTorchGeoNodes allows optimization in the parameter space which has several benefits. From the resulting shapes in this example, it is observable that individual parameters can be scaled independently targeting only specific parts of the shape geometry while preserving the compactness of the 3D shape at the same time.}
    \label{fig:refinement}
\end{figure*}

During a forward pass, we use a hash map that keeps track of the output sockets of individual nodes such that they can be easily accessed by nodes by simply querying the correct hash. Every 'Input node' in the graph parses named parameters, or shape parameters, and initializes the hash map that is updated with each forward call of nodes in the graph. Finally, we accumulate outputs of 'Output nodes' as a list of output geometries. Such design enables flow of gradients as we demonstrate in Figure~\ref{fig:refinement}. Note that in our experiments we only consider computational graphs that have one output but this is not a limitation of our implementation. 

\subsection{Efficient Implementation}

As computational graphs increase in size (in our experiments, graphs can have between 100 and 200 nodes) so does the computational time which is why we considered different ways to improve the efficiency of PyTorchGeoNodes.

Note that, by default, computational nodes are not necessarily pre-sorted in optimal order. A node could request an input that has not been computed yet. As this would lead to several complications and the usage of recursive calls, which would in turn lead to computational bottlenecks, we implement a different solution. During the generation of the computational graph, nodes are sorted into a list using topological sort: Based on dependencies in a graph, we ensure that a node always appears after nodes that it is dependent on. For example, 'Input nodes' and other nodes that do not require any inputs will appear first, and 'Output nodes' will be the last nodes after sorting. Then during inference, we can simply iterate this list, as inputs for nodes will be readily available once the forward pass of the nodes is invoked.

In addition, we observed that some nodes in a graph do not necessarily depend on any inputs. For example, instantiating an initial geometric primitive does not necessarily depend on any parameters. In this case, we implement caching such that the output of such nodes does not need to be recomputed during every forward pass. 
\section{Search Algorithm -- Implementation Details}
\label{sec:search}

\textvars{CD,PC,Sh}

In this section, we discuss implementation details regarding our search algorithm.

\textbf{Discretizing continuous parameters.} Our genetic algorithm, as well as the coordinate descent baseline, rely on discretization of values for continuous parameters during the search. For a given valid range of a parameter, we generate discrete values in linear steps of $0.2$. This value is fitting for all parameters since they are represented in meters, or as aspect ratio relative to other parameters. In the case of object rotation, we use discrete steps of size $90^{\circ}$.

\textbf{Search details.} As discussed in the main paper, our genetic algorithm depends on several different hyper-parameters. Based on empirical observations from our synthetic experiments, we use the following configuration:

\begin{itemize}
    \item Total number of generations is set to 50;
    \item Size of initially randomly generated individuals is $500$. We take $64$ best performing individuals and maintain this population through iterations;
    \item At every iteration we generate $128$ offsprings by randomly selecting pairs of parents and then randomly selecting values for each of the parameters from this pair.
    \item Mutation rate is set to $P_m = 0.9$ and linearly decayed to $0.1$ through generations, and the random noise added for continuous values is $\sigma=0.05$.
    \item To make the search more efficient, we performs refinement every $5$, and we randomly select $20$ percent of offsprings and perform gradient descent until convergence with Adam optimizer and learning rate $0.01$. 
\end{itemize}

\section{Integration of Gaussian Nodes -- Implementation Details}
\label{sec:gaussian}

As explained in the main manuscript, we integrate Gaussian nodes into PyTorchGeoNodes in a straight-forward way, by adjusting the nodes that generate primitive meshes, in our case Cubes and Cylinders, to also generate Gaussian parameters that are fitted to the meshes of corresponding primitives. More precisely, we are inspired by findings from SuGaR~\cite{guedon2024sugar} in order to ensure that object mesh and Gaussians remain consistent:

\begin{itemize}
    \item Means are not optimized directly, we compute means explicitly from triangles of object mesh, and we use $4$ means per face in barycentric coordinate system relative to the individual faces. Instead, we optimize offsets of vertices of the object mesh. 
    \item We keep Gaussians flat by frozing their thickness to $10^{-3}$, and only optimize the remaining two scales. They are initialized to be isotropic and are set to $0.5$ of maximum side length of corresponding triangles. We optimize offsets to these scales.
    \item Gaussian rotations are initialized to be aligned with the object mesh and we only allow optimization of offset to the in-plane rotation.
    \item We initialize colors from the closest points in the point cloud of the corresponding object.
    \item We also optimize opacity which is initially set to $0.7$ in our experiments.
\end{itemize}

We train the parameters for $1000$ iterations. We use the Adam optimizer, and as is the case for different implementations of Gaussian splats, we observed that using different learning rates for different parameters improves convergence: for mesh vertex offset we use learning rate $10^{-4}$, for scales offsets $5^{-3}$, for colors $10^{-2}$, and for opacity $10^{-2}$.
\section{Shape Program Designs for Validation}
\label{sec:designs}

Here, we provide more details on our designs of individual shape programs.

\textbf{'Chair'} consists of 23 parameters, 19 continuous, and 4 boolean parameters:
\begin{itemize}
    \item 'Legs Type' is a boolean parameter that determines whether a chair has four legs or one star-shaped leg in the middle.
    \item 'Legs Size' is a continuous parameter that determines the thickness of a chair in meters. The valid range of values is in $[0.02, 0.08]$;
    \item 'Has Leg Support' is a boolean parameter that controls whether legs of a four-legged chair are connected with support elements;
    \item 'Support Offset-1' is a continuous parameter that controls the height offset of left and right leg support relative to the seat height. The valid range of values is in $[0, 0.5]$;
    \item 'Support Offset-2' is a continuous parameter that controls the height offset of back and front leg support relative to the seat height. The valid range of values is in $[0, 0.5]$;
    \item 'Bottom Thickness' is a continuous parameter that controls the thickness of the bottom surfaces of one-legged chairs in meters. The valid range of values is in $[0.02, 0.08]$;
    \item 'Bottom Size Scale' is a continuous parameter that controls the radius of the bottom surface relative to seat width. The valid range is in $[0.7, 1.0]$;
    \item 'Star Rotation' is a continuous parameter that controls the rotation of the bottom component of one-legged chairs normals to range $[0,1]$ in radians. The valid range of values is in $[0.0, 1.00]$;
    \item 'Seat Height', 'Seat Width', 'Seat Depth', 'Seat Thickness' are continuous parameters that control the \stefan{geometry of the} seat in meters. The valid ranges are $[0.3, 0.9]$, $[0.4, 0.8]$, $[0.4, 0.6]$ and $[0.04, 0.1]$, respectively;
    \item 'Has Back' is a boolean parameter that controls whether a chair has back elements;
    \item 'Back Height' is a continuous parameter that controls the height of the chair back in meters. The valid range is in $[0.3, 1.0]$;
    \item 'Backrest Scale' is a continuous parameter that scales length of backrest relative to the height of the backrest. The valid range is in $[0.1, 1.0]$;
    \item 'Back Thickness' is a continuous parameter that controls the thickness of the back elements in meters. The valid range is in $[0.02, 0.08]$;
    \item 'Backrest Offset Scale' is a continuous parameter that positions the backrest relative to the height of the back. The valid range is in $[0.0, 1.0]$;
    \item 'Has Arms' is a boolean parameter that controls whether a chair has arms;
    \item 'Arm Depth Scale' is a continuous parameter that controls the arm depth relative to the seat depth. The valid range is in $[0.5,0.8]$;
    \item 'Arm Height' is a continuous parameter that controls the height of arms in meters. The valid range is in $[0.1, 0.3]$;
    \item 'Arm Width' is a continuous parameter that controls the width of arms in meters. The valid range is in $[0.08, 0.15]$;
    \item 'Arm Thickness' is a continuous parameter that controls the thickness of arms in meters. The valid range is in $[0.02, 0.05]$.
\end{itemize}

\textbf{'Sofa'} consists of 19 parameters, 13 continuous parameters and 6 boolean parameters:
\begin{itemize}
    \item 'Width', 'Height', 'Depth' are continuous parameters that control width, height and depth of the base of a sofa in meters. The valid ranges are in $[0.5,2.7]$, $[0.3,0.6]$ and $[0.3, 0.6]$, respectively;
    \item 'Has Legs' is a boolean parameter that controls whether the sofa has legs;
    \item 'Leg Size' is a continuous parameter that controls the size of legs in meters. The valid range is in $[0.03, 0.1]$;
    \item 'Leg Height' is a continuous parameter that controls the height of legs in meters. The valid range is in $[0.01, 0.2]$;
    \item 'Has Left Arm' and 'Has Right Arm' are boolean parameters that control whether the sofa has arms;
    \item 'Arm Width', 'Arm Height', 'Arm Depth' are continuous parameters that control the width, height, and depth of arms in meters. The valid ranges are in $[0.05, 0.3], [0.5,0.8], [0.6,1.0]$;
    \item 'Has Arm Legs' is a boolean parameter that controls whether a sofa has legs directly under the arms;
    \item 'Has Back' is a boolean parameter that controls whether a sofa has a back;
    \item 'Back Height' and 'Back Depth' are continuous parameters that control the height and depth of the back in meters. The valid ranges are in $[0.3,0.7]$ and $[0.05,0.3]$;
    \item 'Back Over-Width Scale' determines aspect ratio between width of backrest and seat, and relative to the armrest width. The valid range is in $[0.0,1.0]$;
    \item 'Is L-Shaped' is a boolean parameter that controls whether the sofa contains the 'L' extension;
    \item 'L Width' and 'L Depth' control the width and depth of the 'L' extension in meters. The valid ranges are in $[0.3,0.5], [0.3,1.0]$;
    \item 'Flip L Around Y' is a boolean parameter that controls whether the 'L' extension is on the left or the right side of the couch.
\end{itemize}

\textbf{'Table'} consists of 15 parameters, 11 continuous, and 4 integer parameters:
\begin{itemize}
    \item 'Width', 'Depth' and 'Height' are continuous parameters that control the width, depth, and height of a table in meters. The valid ranges are in $[0.4, 4.0],[0.4,1.3],[0.4,1.5]$;
    \item 'Top' is a boolean that controls whether the shape of the top board of a table is cylindrical or cuboidal;
    \item 'Top thickness' is a continuous parameter that controls the thickness of the top in meters. The valid range is in $[0.04,0.1]$;
    \item 'Legs Type' is an integer parameter that controls whether a table has 1 leg in the middle, or 4 legs on the side.
    \item 'Mid Leg X Scale' and 'Mid Leg Y Scale' are continuous parameters that scale the middle leg relative to the width and depth of the table. The valid range for both parameters is in $[0.05,1.0]$;
    \item 'Has Mid Board' is a boolean parameter that controls whether a table has a second board underneath the top;
    \item 'Mid Board Z Scale' is a continuous parameter that controls the offset of the middle board relative to the height of the table. The valid range is in $[0.05, 0.5]$.
    \item 'Has Cabinet Leg' is a boolean parameter that determines if a four-legged table should have an integrated cabinet in place of legs on one side of the table.
    \item 'Legs Scale X' is a continuous parameter that scales leg of the table relatively to the table width. The valid range is in $[0.01, 0.1]$.
    \item 'Legs Scale Y' is a continuous parameter that scales leg of the table relatively to the table depth. The valid range is in $[0.01, 0.5]$.
    \item 'Legs Offset X' is a continuous parameter that offsets leg of the table relatively to the table width. The valid range is in $[0.0, 1.0]$.
    \item 'Legs Offset Y' is a continuous parameter that offsets leg of the table relatively to the table depth. The valid range is in $[0.0, 1.0]$.
\end{itemize}

\textbf{'Cabinet'} consists of 12 parameters, 8 continuous, 3 boolean, and 1 integer parameter:
\begin{itemize}
    \item 'Width', 'Height', and 'Depth' are continuous parameters that control the width\stefan{, height and depth} of a cabinet in meters. The valid ranges of values are $[0.3, 2.0]$, $[0.3, 2.5]$ and $[0.1,0.6]$, respectively;
    \item 'Board Thickness' is a continuous parameter that controls the thickness of side boards. The valid range of values is $[0.01, 0.09]$;
    \item 'Has Back' is a boolean parameter that controls whether a cabinet has a back board; 
    \item 'Has Legs' is a boolean parameter that controls whether a cabinet has legs;
    \item 'Leg Width', 'Leg Height', 'Leg Depth' are continuous parameters that control the width, height, and depth of legs in meters. The valid range is $[0.03, 0.1]$.
    \item 'Number of Dividing Boards' is an integer parameter that controls the number of dividing boards on a cabinet. The valid range of values is in $[2,5]$;
    \item 'Dividing Board Thickness' is a continuous parameter that controls the thickness of dividing boards in meters. The valid range of values is in $[0.01, 0.05]$;
    \item 'Has Drawers' is a boolean parameter that controls \stefan{whether the cabinet has drawers.}
\end{itemize}

{
    \small
    \bibliographystyle{ieeenat_fullname}
    \bibliography{main}
}
